\documentclass[11pt]{article}

\usepackage{acl}

\usepackage{times}
\usepackage{latexsym}
\usepackage[T1]{fontenc}
\usepackage[utf8]{inputenc}
\usepackage{microtype}
\usepackage{inconsolata}
\usepackage{graphicx}
\usepackage{amsmath,amssymb}
\usepackage{multirow}
\usepackage{enumitem}
\usepackage{xcolor} 
\usepackage{pifont}
\usepackage{makecell}
\newcommand{\cmark}{\textcolor{green!60!black}{\ding{51}}}
\newcommand{\xmark}{\textcolor{red!80!black}{\ding{55}}}
\usepackage{placeins} 
\usepackage{booktabs}
\usepackage{tikz}
\usepackage{pgfplots}
\usepackage{float}
\usepackage[normalem]{ulem} 
\newcommand{\Description}[1]{} 

\title{Drive-P2D: A Progressive Perception-to-Decision Benchmark for VLMs in Autonomous Driving}
\author{
  \textbf{Zecong Tang}$^{1,*}$\;
  \textbf{Zixu Wang}$^{1,*}$\;
  \textbf{Yifei Wang}$^{1,*}$\;
  \textbf{Weitong Lian}$^{1,*}$\\
  \textbf{Tianjian Gao}$^{1}$\;
  \textbf{Haoran Li}$^{1}$\;
  \textbf{Tengju Ru}$^{1}$\;
  \textbf{Lingyi Meng}$^{1}$\\
  \textbf{Zhejun Cui}$^{1}$\;
  \textbf{Yichen Zhu}$^{1}$\;
  \textbf{Qi Kang}$^{1}$\;
  \textbf{Kaixuan Wang}$^{2}$\;
  \textbf{Yu Zhang}$^{1,\dagger}$\\
  \\
  $^{1}$Zhejiang University, Hangzhou, China \\
  $^{2}$The University of Hong Kong, Hong Kong, China\\
  $^{*}$Equal contribution \quad
  $^{\dagger}$Corresponding author
}

\begin{document}
\maketitle

\begin{figure*}[!t]
  \centering
  \includegraphics[width=0.95\linewidth]{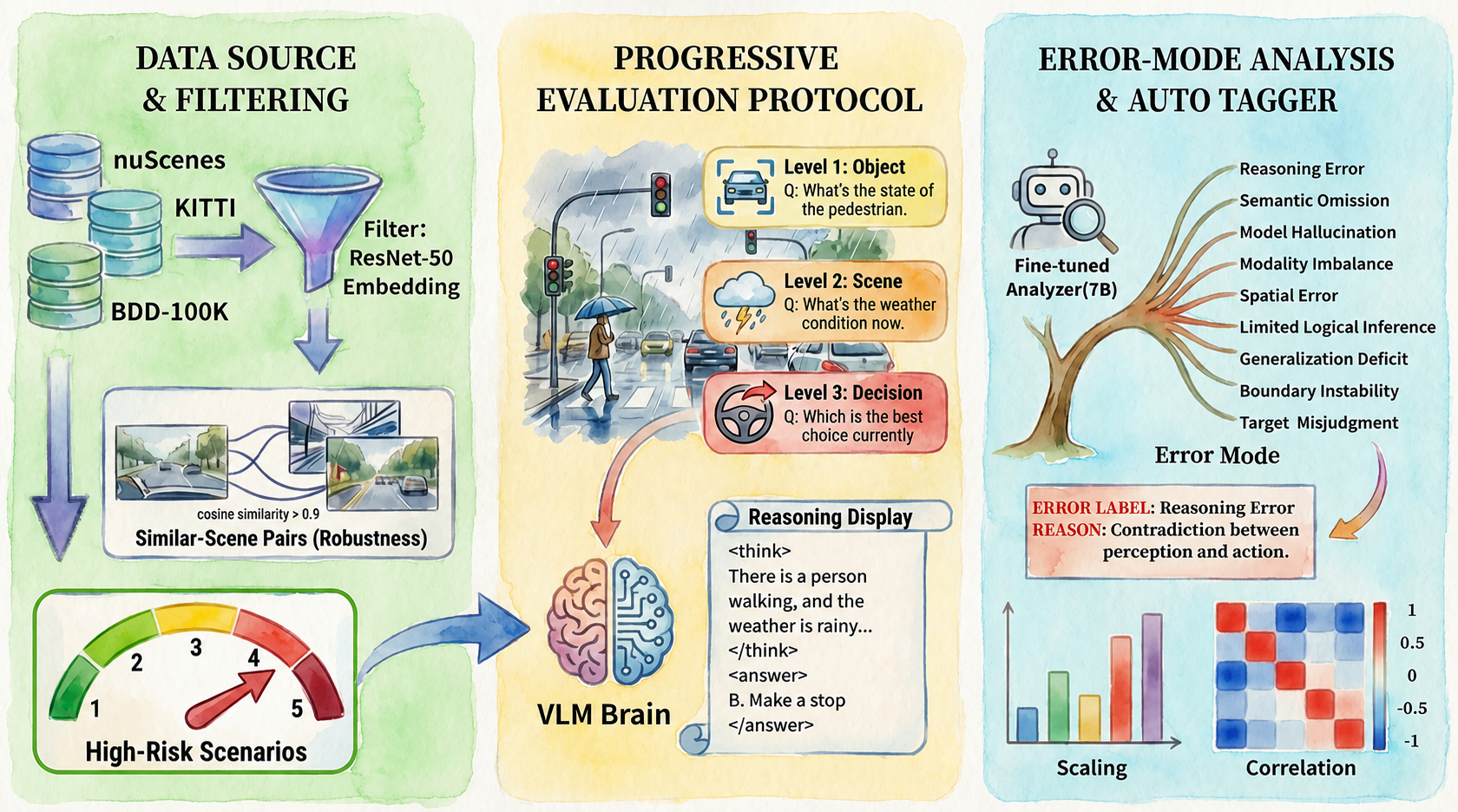}
  \caption{Overview of Drive-P2D. The framework is organized into three stages: (Left) Data Source \& Filtering, (Middle) Progressive Evaluation Protocol, and (Right) Error-Mode Analysis \& Automated Tagging.}
  \Description{Overview of Drive-P2D}
  \label{fig:1}
\end{figure*}

\begin{abstract}
Autonomous driving requires reliable perception and safe decision-making in complex scenarios.
Recent vision-language models (VLMs) demonstrate reasoning and generalization abilities, opening new possibilities for autonomous driving; however, existing benchmarks often evaluate perception and decision-making separately, limit failure analysis with choice-only formats, or introduce evaluation bias through LLM-scored long-form outputs.
To address these issues, we present \textbf{Drive-P2D}, a progressive perception-to-decision benchmark with 6,650 questions across Object, Scene, and Decision levels.
Drive-P2D adopts a separated reasoning-and-answer protocol: final answers are scored objectively, while reasoning is analyzed to identify error modes exposed along the progressive perception-to-decision chain.
We evaluate mainstream VLMs across all and high-risk scenarios, and further characterize the perception-to-decision capability boundary through correlation analysis and similar-scene robustness testing.
Reasoning further exposes failure modes such as logical reasoning errors and semantic feature omissions, and we train a lightweight analyzer model to automate large-scale error-mode annotation of reasoning.
Together, these designs provide practical insights for building safer and more reliable VLMs for real-world autonomous driving.
\end{abstract}


\section{Introduction}

Autonomous driving is a challenging domain operating under diverse and safety-critical conditions.
Research follows two paradigms.
The first, a modular pipeline, separates perception, prediction, and control \citep{paden_survey_2016,schwarting_planning_2018,badue_self-driving_2021,li2022bevformer}.
The second, end-to-end learning, jointly optimizes perception, prediction, planning, and control toward final driving behavior \citep{codevilla_end--end_2018,jiang2023vad,liao2025diffusiondrive}.
While both approaches have achieved notable success, modular systems remain brittle due to error propagation, whereas end-to-end methods often lack robust reasoning about causal structure \citep{badue_self-driving_2021,chib_recent_2023,hu2023planning,chen_end--end_2024}.

The development of large language models (LLMs) has enabled significant advances in instruction following and reasoning \citep{openai_gpt-4_2024,touvron_Llama_2023,deepseek-ai_deepseek-v3_2025}.
In parallel, vision–language models (VLMs) have emerged with capabilities in zero-shot transfer and language-based reasoning \citep{zhang_vision-language_2024,shen_aligning_2024,lu_internvl-x_2025}.
These properties highlight VLMs' potential for autonomous driving; related VLM-based systems have been proposed \citep{zhou_vision_2024,tian_drivevlm_2024,fu_drivegenvlm_2024,you_v2x-vlm_2025}, demonstrating increasing research activity and diverse approaches.

Benchmarks serve as a bridge between methodological advances and practical deployment by revealing model capabilities and failures.
First, many autonomous-driving VLM benchmarks evaluate perception and decision-making with separate metrics, rather than modeling their progressive relation \citep{qian2024nuscenes,guo2026surds,tian2024drivevlm,sima2024drivelm,xie2025vlms}.
Second, some benchmarks use choice-only formats, which enable objective scoring but provide limited evidence for diagnosing why models fail \citep{qian2024nuscenes,guo2026surds}.
Third, other benchmarks require long-form textual outputs and rely on LLM-based scoring, which can introduce evaluator bias and reduce the objectivity of performance comparison \citep{tian2024drivevlm,xie2025vlms}.

To address these limitations, we introduce Drive-P2D (Figure~\ref{fig:1}), a progressive perception-to-decision benchmark for autonomous-driving VLMs.
Drive-P2D is constructed from nuScenes, KITTI, and BDD100K, and organizes tasks into three progressive levels—Object, Scene, and Decision.
The benchmark comprises 6,650 questions across six tasks, with evaluations stratified by scenario risk.
Drive-P2D adopts a separated reasoning-and-answer protocol, where final choice answers are used for objective scoring and reasoning is analyzed to identify problems exposed along the perception-to-decision chain.
Using final choices, we evaluate mainstream VLMs to characterize capability boundaries and quantify perception--decision dependencies through correlation analysis.
To further assess robustness, the benchmark includes 60 pairs of visually similar scenarios.
We also examine InternVL scaling and identify key inflection points.
For reasoning analysis, we collect and annotate model-generated reasoning from multiple VLMs to identify error modes.
We further fine-tune a lightweight analyzer to tag error modes in new reasoning, supporting scalable evaluation and model optimization.
Table~\ref{tab:benchmark-comparison} compares Drive-P2D with representative autonomous-driving VLM benchmarks.

Our work makes three key contributions:

\begin{itemize}[leftmargin=2em, topsep=0pt, itemsep=-2.5pt]
\item We introduce Drive-P2D, a progressive perception-to-decision benchmark with a three-level protocol, covering 6,650 questions with six tasks and risk-aware splits.
\item We evaluate mainstream VLMs via choice-question scoring, analyzing perception--decision dependencies, similar-scene robustness, and InternVL scaling behavior.
\item We analyze reasoning to reveal error modes, and fine-tune a lightweight analyzer model to automate error-mode labeling, enabling large-scale error-mode analysis.
\end{itemize}

\begin{table*}[t]
\centering
\caption{Comparison with representative autonomous-driving VLM benchmarks. P--D Analysis denotes perception--decision analysis; Reasoning Analysis indicates whether model-generated reasoning or reasoning-related scoring and analysis are used; Error Modes indicates whether model-generated reasoning is categorized into defined error modes; Auto Tagger indicates whether automated error-mode tagging is supported.}
\label{tab:benchmark-comparison}
\setlength{\tabcolsep}{4.2pt}
\renewcommand{\arraystretch}{1.00}
\small
\begin{tabular*}{\textwidth}{@{\extracolsep{\fill}}lccccccc@{}}
\toprule
Benchmark &
\makecell[c]{Progressive\\Evaluation} &
\makecell[c]{P--D\\Analysis} &
\makecell[c]{Risk-aware\\Split} &
\makecell[c]{Similar-scene\\Pairs} &
\makecell[c]{Reasoning\\Analysis} &
\makecell[c]{Error\\Modes} &
\makecell[c]{Auto\\Tagger} \\
\midrule
nuScenes-QA~\citep{qian2024nuscenes}  & \xmark & \xmark & \xmark & \xmark & \xmark & \xmark & \xmark \\
SURDS~\citep{guo2026surds}            & \xmark & \xmark & \xmark & \xmark & \xmark & \xmark & \xmark \\
DriveLM~\citep{sima2024drivelm}       & \cmark & \xmark & \xmark & \xmark & \cmark & \xmark & \xmark \\
DriveVLM~\citep{tian2024drivevlm}     & \cmark & \xmark & \xmark & \xmark & \cmark & \xmark & \xmark \\
DriveBench~\citep{xie2025vlms}        & \xmark & \xmark & \xmark & \xmark & \cmark & \xmark & \xmark \\
OmniDrive~\citep{wang2025omnidrive}   & \cmark & \cmark & \xmark & \xmark & \cmark & \xmark & \xmark \\
\midrule
Drive-P2D                              & \cmark & \cmark & \cmark & \cmark & \cmark & \cmark & \cmark \\
\bottomrule
\end{tabular*}
\end{table*}
\section{Related Work}
\subsection{VLMs for Autonomous Driving}

Recent advances in LLMs have accelerated the development of VLMs \cite{touvron_Llama_2023,openai_gpt-4_2024,deepseek-ai_deepseek-v3_2025,zhao2026surveylargelanguagemodels}. 
VLMs leverage large-scale image–text pretraining and instruction tuning, demonstrating robust instruction-following, zero-shot generalization, and contextual reasoning capabilities \cite{li_blip-2_2023,wang_qwen2-vl_2024,lu_internvl-x_2025}. 
They have been extensively investigated in autonomous driving for scene understanding, interactive reasoning, and end-to-end planning \cite{zhou_vision_2024,xu2024drivegpt4,gopalkrishnan_multi-frame_2024,you_v2x-vlm_2025}. 
However, current VLMs still suffer from hallucinations, brittle reasoning, and omissions of critical visual cues in complex scenes \cite{guan_hallusionbench_2024,yu_mm-vet_2024,bai_hallucination_2025}.

\subsection{Benchmarks for LLMs/VLMs}

For LLMs, mainstream benchmark suites evaluate knowledge, reasoning, and math/coding abilities using more challenging protocols and adversarial design \cite{clark_think_2018,sakaguchi_winogrande_2021,chen_evaluating_2021,rein2023gpqa,wang_mmlu-pro_2024,patel_aime_2024}. 
For VLMs, general-purpose evaluations aim to assess comprehensive capabilities across perception, grounding, OCR reasoning, and hallucination robustness \cite{fu2024mmecomprehensiveevaluationbenchmark,liu2024mmbench,yu_mm-vet_2024,yue_mmmu_2024,guan_hallusionbench_2024,liu_ocrbench_2024,yue_mmmu-pro_2025,tang_mtvqa_2025,zhang_vcr_2025}. 
In autonomous driving, recent VLM-oriented benchmarks provide driving scenes and tasks for perception–planning evaluation \cite{guo2024drivemllmbenchmarkspatialunderstanding,li2024automated,li2025benchmarkingassessingsafetyrobustness,sima_drivelm_2025}; however, they often decouple perception from decision-making, constrain failure diagnosis, or rely on potentially biased LLM-based scoring.

\section{Drive-P2D}


\subsection{Data Source and Filtering}
\label{sec:data-source-filtering}
\textbf{Data source.}
We construct our benchmark from three widely used driving datasets—nuScenes, KITTI, and BDD100K \cite{geiger2013vision, caesar2020nuscenes, yu2020bdd100k}—focusing on front-facing images.
The integration of these datasets spans multiple regions and driving environments.
This diversity mitigates dataset-specific bias and enhances the coverage of driving conditions.
Since we attach multiple task-specific questions to the same driving scene, a single image may correspond to multiple questions across tasks; Appendix~\ref{app:datasets-usage} summarizes the dataset profiles and statistics.

\textbf{Data filtering.}
For quality control, we employ a similarity-based filtering procedure.
We compute image embeddings using a pretrained ResNet-50 \cite{he2016deep} with global average pooled representations and construct candidate pairs based on cosine similarity.
Pairs exceeding a similarity threshold of 0.9 are treated as near-duplicate cases and removed to reduce redundancy.

\subsection{Benchmark Construction}
We adopt a three-level protocol, which is progressively structured: object-level perception provides task-relevant cues, scene-level understanding adds contextual constraints, and decision-level tasks evaluate how these cues and constraints support driving decisions.

\textbf{Overview.}
We define two tasks at each level, thereby establishing an Object–Scene–Decision evaluation pipeline.
All tasks are selection questions: Object and Decision are single-choice, while Scene is multiple-choice.
The benchmark contains \textbf{6,650} questions covering the three levels.
Detailed annotation templates for all six tasks are provided in Appendix~\ref{app:annotation-templates}.
We further include risk-aware splits, similar-scene robustness testing, and perception--decision correlation analysis to support fine-grained evaluation beyond overall accuracy.

\textbf{Task levels.}
Object level evaluates perception: (Object-1) identify the object in the image that most influences the decision; (Object-2) determine the state of a designated object.
Scene level aims to evaluate scene-level understanding: (Scene-1) recognize weather and illumination; (Scene-2) identify special scene factors that potentially influence driving decisions (e.g., roadworks, accidents).
Decision level aims to evaluate decision-making capability: (Decision-1) select the optimal action for the ego vehicle in the scenario; (Decision-2) evaluate the safety of specified and potentially suboptimal actions.
Decision-1/2 are designed around perceptual cues and contextual constraints from Object-1/2 and Scene-1/2.
We summarize the design intentions of the six tasks in Appendix~\ref{app:task-intentions}.

\textbf{Annotation protocol.}
To ensure annotation accuracy, all QA pairs are manually annotated by human annotators.
To ensure annotation consistency, each item is independently labeled by two experienced drivers.
In cases of disagreement, a third human arbitrator adjudicates, and the final annotation is assigned accordingly.

\textbf{High-risk scenario design.}
Due to the scarcity of high-risk scenarios, collecting sufficient data for these cases can be challenging and may limit the pretraining of VLMs.
However, these scenarios are critical for safe driving, and thus we evaluate them to assess the generalization capability of VLMs.
Each scenario is rated by two expert annotators using a five-point risk scale (1: minimal, 5: severe, see Appendix~\ref{app:risk-rubric}).
The average score is used as the final rating, and scenarios with a score of 4 or higher are designated as high-risk scenarios, resulting in a total of \textbf{1.6K} questions for analysis.

\textbf{Similar-scene robustness.}
Humans can identify task-relevant information and ignore redundant cues, enabling accurate decision-making in visually similar scenarios.
To test whether models exhibit this fine-grained capability, we construct \textbf{60} near-duplicate pairs based on similarity filtering in Sec.~3.1.
We evaluate model performance on Decision-1 using both the single-image accuracy and the joint accuracy (the probability that both images in a pair are answered correctly).
Comparison with the squared baseline of individual accuracies evaluates whether decisions reflect causal understanding rather than superficial feature associations, thereby assessing VLMs’ robustness in causal reasoning.
Representative examples are provided in Appendix~\ref{app:similar}.

\textbf{Correlation analysis.}
Benefiting from our three-level design (Object, Scene, and Decision), which establishes a progressive rather than independent structure, we explore the internal consistency of model behavior across perception and decision dimensions by computing pairwise Pearson correlations among all task scores for each model.
This analysis reveals whether enhancements in one capability co-occur with gains or trade-offs in others.
The resulting correlation matrices characterize how performance transfers across the progressive perception-to-decision levels.

\begin{figure*}[!t]
  \centering
  \includegraphics[width=0.95\linewidth]{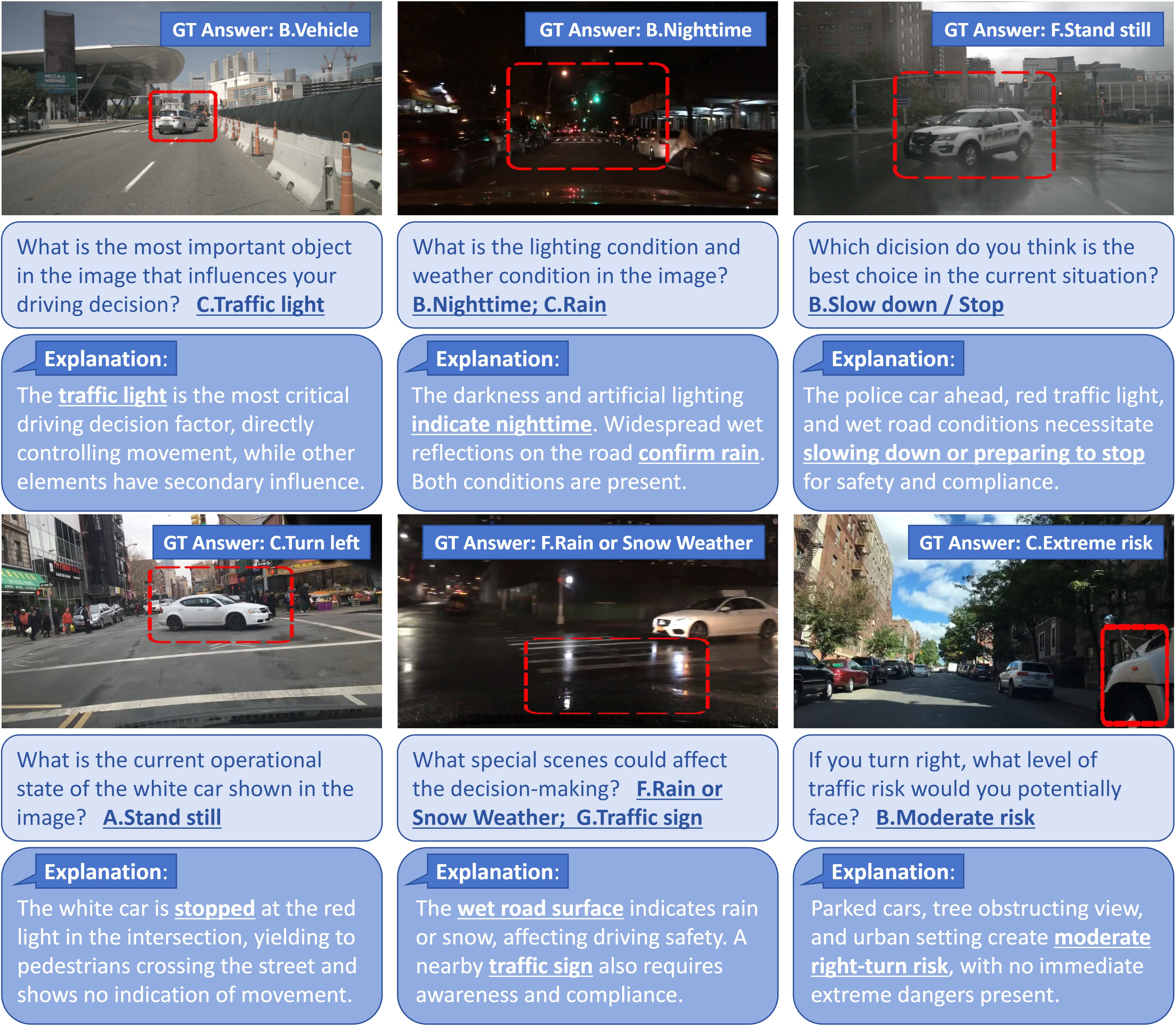}
  \caption{GPT-4.1 Failure Cases. The six subfigures illustrate GPT-4.1 failure cases across different tasks. Each subfigure consists of the given image, the question, the model’s answer, the ground truth, and its reasoning.}
  \label{fig:bad}
\end{figure*}

\begin{table*}[t]
\centering
\caption{Evaluation on All Scenarios (0-shot)}
\label{tab:app-all-0shot}
\setlength{\tabcolsep}{6pt}\renewcommand{\arraystretch}{1.05}\small
\begin{tabular}{lccccccccc}
\toprule
Model & Size & Open & Object-1 & Object-2 & Scene-1 & Scene-2 & Decision-1 & Decision-2 & Avg Score \\
\midrule
Gemini-2.5-Pro & - & \xmark & 62.90 & \textbf{\underline{74.35}} & 76.73 & 25.61 & 49.47 & \textbf{\underline{55.11}} & 57.36 \\
GPT-4.1 & - & \xmark & 67.48 & 73.74 & \textbf{\underline{92.54}} & \textbf{\underline{46.44}} & \textbf{\underline{56.95}} & 51.91 & \textbf{\underline{64.84}} \\
\midrule
Gemma & 4B & \cmark & 49.86 & 50.07 & 27.34 & \phantom{0}0.89 & 45.78 & 30.66 & 34.10 \\
Gemma & 27B & \cmark & 64.46 & 63.55 & 63.45 & \phantom{0}7.20 & 50.40 & 31.65 & 46.78 \\
Llama & 11B & \cmark & 58.67 & 53.45 & 87.00 & 18.37 & 41.84 & 37.99 & 49.55 \\
Llama & 90B & \cmark & 51.39 & 63.48 & 91.75 & 36.78 & 45.14 & 49.55 & 56.35 \\
Llava & 7B & \cmark & 26.23 & 52.95 & 87.15 & 32.49 & 41.40 & 29.87 & 45.01 \\
Llava & 72B & \cmark & 45.88 & 66.72 & 85.66 & 35.42 & 51.60 & 34.89 & 53.36 \\
Phi & 6B & \cmark & 60.23 & 63.10 & 91.01 & 21.67 & 44.58 & 30.72 & 51.89 \\
Qwen & 7B & \cmark & 65.58 & 59.34 & 65.16 & 33.60 & 40.87 & 35.48 & 50.00 \\
Qwen & 72B & \cmark & \textbf{\underline{68.06}} & 68.24 & 77.19 & 37.62 & 54.69 & 38.20 & 57.33 \\
\bottomrule
\end{tabular}
\end{table*}

\subsection{Error-Mode Analysis Settings}\label{sec:error-mode-settings}

We analyze the reasoning process of VLMs by classifying errors and fine-tune a lightweight model to scale up analysis for automated evaluation.

\textbf{Model reasoning.}
Reasoning provides observable diagnostic signals that expose the model's stated analysis, including the evidence it uses, the constraints it considers, and the steps it follows before producing a final answer.
To obtain these signals, we instruct the model to generate reasoning enclosed within \texttt{<think>…</think>}, followed by the final answer enclosed within \texttt{<answer>…</answer>} in a single-turn interaction.
This configuration separates reasoning analysis from answer scoring, enabling fine-grained diagnosis while keeping the final evaluation based on explicit answer correctness.

\textbf{Error categories.}
We define nine error categories (E1–E9) to characterize failures across the decision-making process, following the taxonomy in Appendix~\ref{app:exp-taxonomy}.
The three primary categories are (E1) Logical Reasoning Error (the reasoning chain violates causality), (E2) Semantic Feature Omission (the model overlooks or misjudges semantic/visual cues such as turn signals), and (E3) Model Hallucination (the model invents non-existent objects, attributes, or relations).
The remaining categories are (E4) Modality Imbalance, (E5) Spatial Relation Misjudgment, (E6) Limited Logical Inference, (E7) Generalization Deficit, (E8) Decision Boundary Instability, and (E9) Target Priority Misjudgment.

\textbf{Automated error-mode tagging.}
Manual error-mode analysis is resource-intensive and constrains scalability.
To enable automated large-scale evaluation, we train a lightweight analyzer model with 7B parameters that identifies error modes.
The model takes as input: (1) the image, question, options, and ground-truth answer; (2) the tested model’s reasoning and final answer.
It outputs one or more error-mode labels from the nine error categories, or a no-error label otherwise.
This approach enables scalable error-type identification, and supports systematic evaluation of model reasoning.
The analyzer i/o format, training details and the prompt templates are provided in Appendix~\ref{app:exp-io}.

\section{Experiments}

\subsection{All Scenarios Results}\label{sec:all-0shot}
We analyze the overall results to evaluate general capability.
GPT-4.1 ranks first in overall performance, while among open-source models, Qwen (72B) yields the highest.
Although closed-source models show strong overall performance, recent open-source systems show growing competitiveness.
Representative GPT-4.1 failure cases are shown in Figure~\ref{fig:bad}, complete 0-shot results in Table~\ref{tab:app-all-0shot}, and the experimental setup in Appendix~\ref{app:exper_setup}.

The results show clear stratification: larger models generally perform better on Object-1/2 and more consistently on Scene-1, while models under 10B show higher variance across tasks, highlighting the role of scale in multi-task stability.
On Scene-2, most models exhibit substantially lower accuracy, suggesting that current VLMs struggle with robust scene understanding and often fail to accurately interpret special scene factors that should constrain downstream decisions.
For Decision-1/2, most VLMs still lack the stability observed in perception tasks and exhibit high variability under complex conditions, suggesting that decision-making is not sufficiently supported by upstream object and scene-level information, especially when scene understanding remains difficult.

\subsection{High-Risk Scenario Results}\label{sec:highrisk-0shot}
We further analyze high-risk scenarios under the 0-shot setting.
Although high-risk scenarios are scarce, the average scores are higher than those on all scenarios, with improvements mainly concentrated in decision tasks among leading models.
Meanwhile, compared with all scenarios, object-level recognition tends to improve under high-risk conditions, but Scene-2 remains a significant challenge across models.
Complete 0-shot high-risk results are reported in Appendix~\ref{app:highrisk-0shot}.

Relative to all scenarios, high-risk scenarios further widen the performance gap between large and small models.
This suggests that hazardous scenes impose a more challenging test of model generalization.
Large models exhibit more robust decision-making because they better preserve salient object cues and scene-level risk constraints when forming safety-oriented decisions.
In contrast, small models fail to exploit these upstream cues and show limited cross-level integration from perception and scene understanding to decision-making.
Although large models improve under high-risk conditions, current capabilities remain insufficient for direct deployment in safety-critical settings.

\begin{figure*}[!t]
  \centering
  \includegraphics[width=\linewidth]{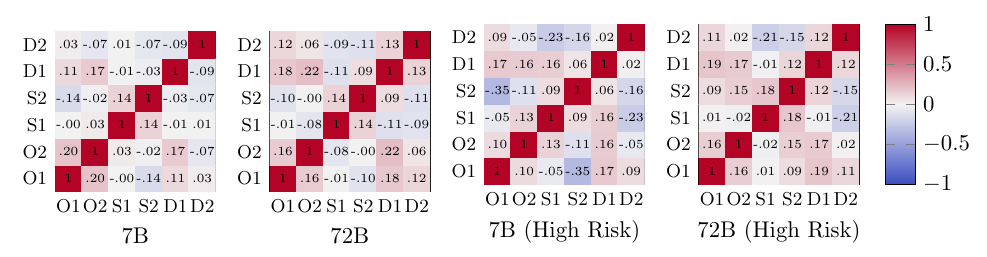}
  \caption{Correlation Matrices of Qwen (7B) and Qwen (72B)}
  \Description{Correlation matrix for Qwen2.5-VL-Instruct}
  \label{fig:3}
\end{figure*}

\subsection{Few-Shot Learning Results}\label{sec:fewshot-all}

We analyze the impact of few-shot prompting on overall performance and individual task dimensions.
Considering average performance, some models exhibit small improvements, whereas others exhibit performance decline or a non-monotonic trend, with initial declines at 1-shot followed by partial recovery at higher shot counts.
Performance gains are predominantly observed in larger models, while smaller models generally show a performance decline or variability.
At the task level, certain tasks maintain relatively high performance, whereas tasks requiring cross-level integration—such as Scene-2 and Decision-2—tend to lead to more errors and may result in substantial performance degradation.
An overview of all-scenario / high-risk trends under 0/1/2/5-shot prompting is shown in Appendix~\ref{app:radar-charts} (Figures~\ref{fig:app-all-radar} and~\ref{fig:high-risk-all-radar}); comprehensive per-shot results are summarized in Appendix Table~\ref{tab:app-eval-intern-excluded-1} and~\ref{tab:app-eval-intern-excluded-2}.

Few-shot prompting produces different outcomes across models, indicating that decision-making under uncertainty may not consistently benefit from examples.
A small number of examples can provide task priors and response patterns, but may also bias tasks such as Decision-2 toward example priors when alignment is fragile.
This suggests that few-shot performance depends not only on the number of examples, but also on whether the examples provide reliable upstream cues and decision criteria for cross-level tasks.

\subsection{Perception–Decision Correlation Analysis}\label{sec:correlation}

To investigate the internal consistency of model behavior across levels, we computed Pearson correlation coefficients between Object (O), Scene (S), and Decision (D) dimensions under both all and high-risk scenarios.
Figure~\ref{fig:3} shows correlation matrices for Qwen-7B and -72B.
Extended correlation matrices covering all prompt settings and scenario types are provided in Appendix~\ref{app:corr-qwen-total}.

Correlations across tasks are generally weak (mostly $-0.2$ to $0.2$), with only Decision-1/2 showing mild positive associations with Object recognition, suggesting limited alignment between object-level cues and decision performance.
Across both settings, 7B shows weaker cross-level correlations than 72B, with this weakness amplified in high-risk scenarios (e.g., Scene--Decision down to $r=-0.23$).
Although 72B remains more balanced in high-risk scenarios, the overall correlations remain weak, suggesting that better upstream performance does not reliably transfer to decision-making.
Overall, while scaling improves stability and reduces negative task interactions, it does not substantially strengthen cross-task coupling, suggesting that tighter perception--decision integration may require changes beyond model scale.

\subsection{Robustness under Similar-Scene Pairs}

For each model, we evaluate accuracy on individual decision tasks (Decision-1) and the probability that both images in a closely related pair are answered correctly (Decision-1 (both)).
If a model relies on task-relevant upstream cues when making decisions, the joint accuracy should approximate the product of the individual accuracies; substantially lower joint performance suggests that shared but irrelevant cues influence decision-making.
Detailed results are provided in Appendix~\ref{app:similar-scene}.

Analysis reveals the following patterns: For almost all models, the joint correct rate is lower than the expected squared baseline.
Models such as Llama (11B) and Llava (7B) show significant drops ($p<0.05$ or $p<0.10$).
This pattern suggests that small-scale models tend to recombine perceptual elements without reliably identifying which upstream visual cues are decision-relevant.
Larger models demonstrate limited robustness, with Llava (72B) achieving the highest performance on both individual and paired tests.

The pronounced gap between individual and joint accuracies, particularly in smaller models, indicates that most systems fail to identify the objects or scenes that are consistently decision-relevant.
This is consistent with reliance on superficial compositional patterns rather than stable decision-relevant semantics.
Overall, current models still struggle to suppress irrelevant perceptual priors and to maintain reliable perception-to-decision transfer in visually similar scenarios.

\begin{figure}[t]
  \centering
  \includegraphics[width=1.0\linewidth]{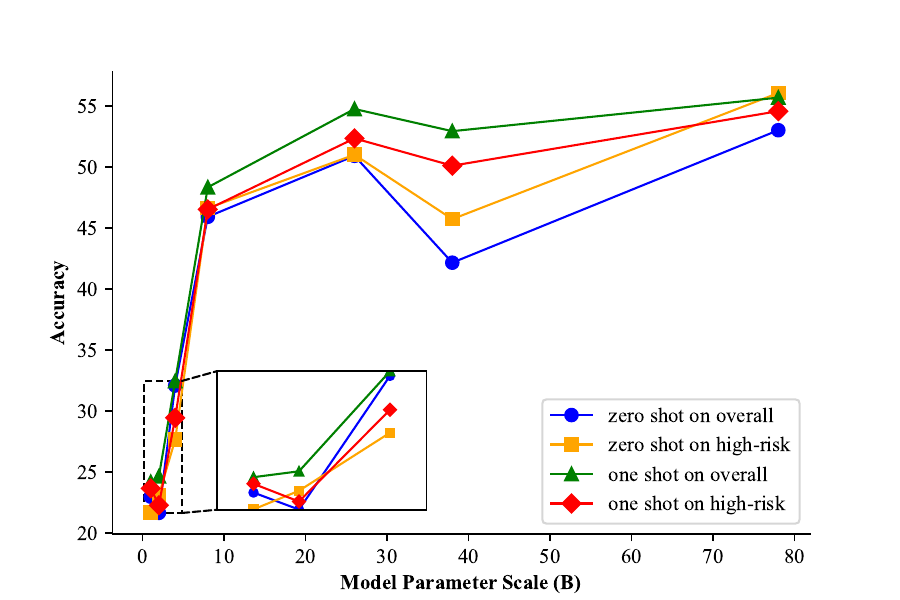}
  \caption{Scaling behavior of InternVL models}
  \Description{###}
  \label{fig:5}
\end{figure}

\begin{figure*}[t]
  \centering
  \includegraphics[width=0.95\textwidth]{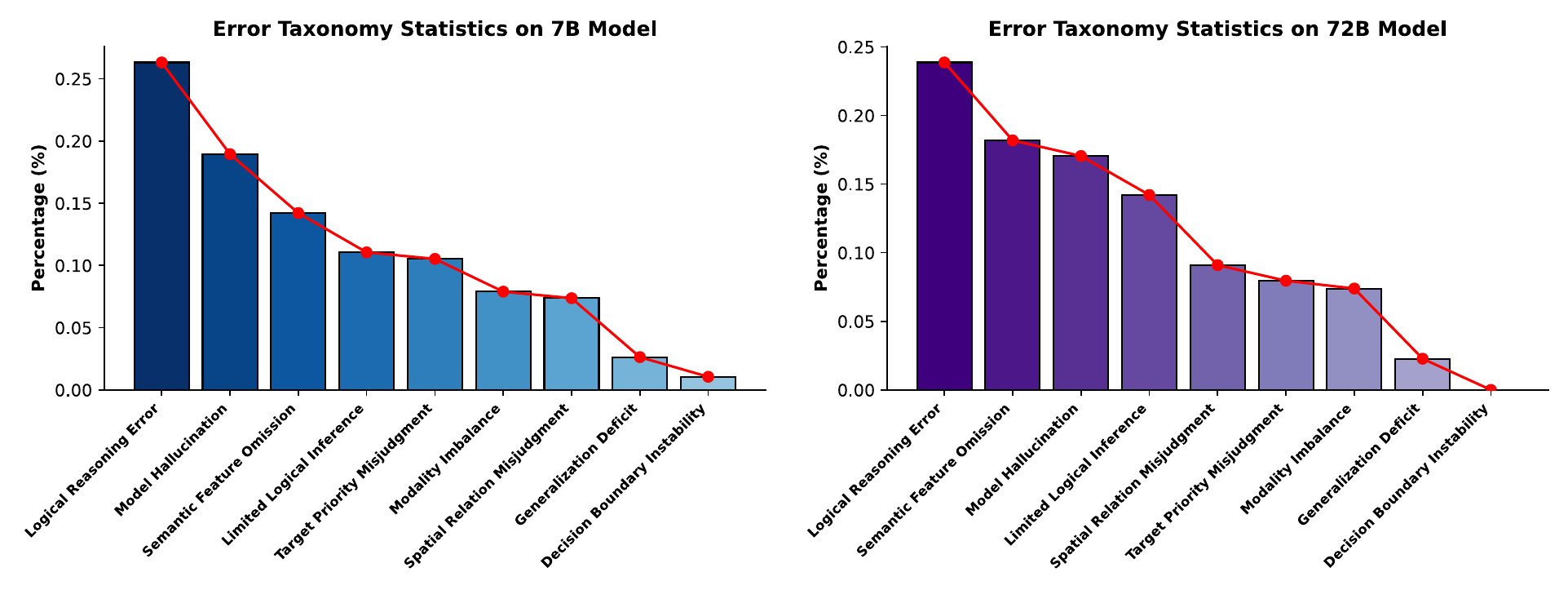}
  \caption{Results of Error-Mode Analysis for Qwen (7B) and Qwen (72B)}
  \Description{###}
  \label{fig:4}
\end{figure*}

\begin{table*}[t]
\centering
\caption{Performance of Analyzer Model and Baselines (\%)}
\label{tab:exp-match}
\setlength{\tabcolsep}{6pt}\renewcommand{\arraystretch}{1.05}\small
\begin{tabular}{lcccccc}
\toprule
Model & Size & Open & Exact Match $\uparrow$ & Partial Match & Mismatch $\downarrow$ & Avg Score $\uparrow$ \\
\midrule
GPT-4.1       & - & \xmark             & 48.09 & 23.02 & 28.88 & 59.60 \\
Qwen          & 7B & \cmark & 48.09 & 14.39 & 37.52 & 55.29 \\
Qwen          & 72B & \cmark & 50.25 & 21.58 & 28.17 & 61.04 \\
Llama         & 11B & \cmark & 17.16 & 20.86 & 61.98 & 27.59 \\
Llama         & 90B & \cmark & 40.18 & 25.18 & 34.64 & 52.77 \\
Ours          & 7B & \cmark & \textbf{\underline{65.36}} & 11.51 & \textbf{\underline{23.13}} & \textbf{\underline{71.11}} \\
\bottomrule
\end{tabular}
\end{table*}

\subsection{Validation of the ``Scaling Behavior''}\label{sec:scaling}

We evaluate the scaling behavior of parameter size in the InternVL family.
In traditional scaling laws, model performance increases with parameter size and typically follows an approximately logarithmic trend; however, on our benchmark, the 38B model exhibits a pronounced performance drop (Figure~\ref{fig:5}).
This anomaly persists under both 0-/1-shot settings and across all-scenario and high-risk splits.

The 38B model exhibits deviations in its reasoning: compared with smaller models, it produces more step-by-step stated analysis, but relative to the larger model, its reasoning remains incomplete.
For instance, in the Scene-2 task, the 38B model categorizes options by factors such as lighting and weather conditions, but assumes that one option must be selected from each category, even when no correct answer exists.
This excessively constrained stated analysis can lead to decision-making errors and performance degradation.

These observations have important implications for VLM evaluation.
Model performance does not scale monotonically with parameter size, as intermediate model sizes may exhibit a mismatch between reasoning quality and task demands.
Moreover, combining answer-based scores with error-mode analysis of reasoning helps identify where the progressive chain breaks down, rather than only reporting non-monotonic performance trends.

\subsection{Error-Mode Analysis Results}\label{sec:error-mode-analysis}

We conducted error-mode analysis on the Qwen-7B and Qwen-72B models.
Results are presented in Figure~\ref{fig:4}.
The results indicate that model reasoning often contains varying degrees of flaws.
Even when some final answers are correct, the corresponding reasoning can still be inconsistent or erroneous.

Figure~\ref{fig:4} reveals that while the relative ranking of error-weight distributions slightly differs between models of varying sizes, logical reasoning errors, semantic feature omissions, and model hallucinations consistently occupy the top three error categories—with logical reasoning errors being predominant, indicating that current VLMs still struggle to produce coherent stated analysis for multimodal driving decisions.

The error-mode annotations further reveal boundary-related failures, such as unstable distinctions between lane changes and turns, showing that VLMs struggle to maintain consistent decision criteria in ambiguous driving situations.
Overall, the error distribution suggests that current VLMs still struggle to produce reliable analysis, including using relevant visual evidence, integrating scene constraints, and making consistent action judgments, highlighting the value of reasoning analysis.

\subsection{Evaluation Results of Analyzer Model}

For model training and evaluation, we collected 1,680 reasoning outputs from the previously assessed models, tasks, and all shot settings.
1,500 reasoning outputs were used for training, while 180 reasoning outputs were reserved for evaluation.
We employ Qwen (7B) as the base model, and perform supervised fine-tuning to classify error modes in reasoning according to predefined categories.

We evaluate the fine-tuned 7B analyzer against mainstream baselines, as summarized in Table~\ref{tab:exp-match}.
Our model achieves the highest Exact Match and the lowest Mismatch, demonstrating stronger label agreement than competitive baselines.
The Average Score is also the best, indicating consistent label agreement across diverse cases.
Consequently, the analyzer's outputs can support large-scale auxiliary assessment and automated labeling.
Appendix~\ref{app:analyzer-robustness} further evaluates analyzer stability on an expanded 480-output set across different model sources and high-risk and non-high-risk subsets.

\section{Conclusion}
This paper presents Drive-P2D, a progressive perception-to-decision benchmark for evaluating VLMs in autonomous driving scenarios, using a separated reasoning-and-answer protocol.
Through experiments on multiple VLMs, we draw three main conclusions.
(1) VLMs perform relatively better on object-level perception, while scene understanding and decision-making remain weaker; although leading models improve on high-risk decision tasks, performance is still insufficient for safety-critical deployment.
(2) This perceptual advantage does not reliably transfer to decision-making, as reflected by weak perception--decision correlations and limited robustness under similar scenarios.
(3) Reasoning reveals failure modes in the model's stated analysis, such as logical reasoning errors, semantic feature omissions and other problems; our lightweight analyzer model further enables scalable automated error-mode tagging.
This benchmark provides a systematic foundation for studying the capability boundaries and reasoning patterns of VLMs, and offers insights for building more reliable autonomous driving systems.

\section*{Limitations}
Our study, benchmark, and analyses have several limitations.%
(1) Data scope: Drive-P2D is constructed from front-facing camera images in nuScenes, KITTI, and BDD100K; it does not include multi-camera setups, multi-sensor inputs (e.g., LiDAR, radar), or video temporal information, which constrains temporal reasoning and occlusion handling. (2) Task format: We employ only single-/multiple-choice questions to reduce ambiguity and enable objective scoring, which may underestimate models capable of producing richer free-form reasoning outputs and may induce answer priors. (3) Model coverage: We evaluate representative open- and closed-source models as well as InternVL scales, yet vendor updates and inference settings (e.g., temperature, generation length) can affect the results.%

\section*{Ethical Considerations}
We use only public datasets under their research licenses. The data may contain pedestrians, vehicles, and buildings in public spaces; no attempts were made to identify individuals, and no new personal data were collected. Annotators were team members informed of the task requirements. The benchmark is intended solely for offline research and is not to be used to determine the roadworthiness of any driving system; prominent disclaimers will be provided. To mitigate misuse risks, such as selective citation or overfitting, we will release the full protocol and open-source the relevant code.


\FloatBarrier
\bibliography{ref}

@misc{fu2024mmecomprehensiveevaluationbenchmark,
      title={MME: A Comprehensive Evaluation Benchmark for Multimodal Large Language Models}, 
      author={Chaoyou Fu and Peixian Chen and Yunhang Shen and Yulei Qin and Mengdan Zhang and Xu Lin and Jinrui Yang and Xiawu Zheng and Ke Li and Xing Sun and Yunsheng Wu and Rongrong Ji},
      year={2024},
      eprint={2306.13394},
      archivePrefix={arXiv},
      primaryClass={cs.CV},
      url={https://arxiv.org/abs/2306.13394}, 
}

@inproceedings{liu2024mmbench,
  title={Mmbench: Is your multi-modal model an all-around player?},
  author={Liu, Yuan and Duan, Haodong and Zhang, Yuanhan and Li, Bo and Zhang, Songyang and Zhao, Wangbo and Yuan, Yike and Wang, Jiaqi and He, Conghui and Liu, Ziwei and others},
  booktitle={European conference on computer vision},
  pages={216--233},
  year={2024},
  organization={Springer}
}

@inproceedings{yu2020bdd100k,
  title={Bdd100k: A diverse driving dataset for heterogeneous multitask learning},
  author={Yu, Fisher and Chen, Haofeng and Wang, Xin and Xian, Wenqi and Chen, Yingying and Liu, Fangchen and Madhavan, Vashisht and Darrell, Trevor},
  booktitle={Proceedings of the IEEE/CVF conference on computer vision and pattern recognition},
  pages={2636--2645},
  year={2020}
}

@inproceedings{caesar2020nuscenes,
  title={nuscenes: A multimodal dataset for autonomous driving},
  author={Caesar, Holger and Bankiti, Varun and Lang, Alex H and Vora, Sourabh and Liong, Venice Erin and Xu, Qiang and Krishnan, Anush and Pan, Yu and Baldan, Giancarlo and Beijbom, Oscar},
  booktitle={Proceedings of the IEEE/CVF conference on computer vision and pattern recognition},
  pages={11621--11631},
  year={2020}
}

@inproceedings{sima2024drivelm,
  title={Drivelm: Driving with graph visual question answering},
  author={Sima, Chonghao and Renz, Katrin and Chitta, Kashyap and Chen, Li and Zhang, Hanxue and Xie, Chengen and Bei{\ss}wenger, Jens and Luo, Ping and Geiger, Andreas and Li, Hongyang},
  booktitle={European Conference on Computer Vision},
  pages={256--274},
  year={2024},
  organization={Springer}
}

@article{geiger2013vision,
  title={Vision meets robotics: The kitti dataset},
  author={Geiger, Andreas and Lenz, Philip and Stiller, Christoph and Urtasun, Raquel},
  journal={The international journal of robotics research},
  volume={32},
  number={11},
  pages={1231--1237},
  year={2013},
  publisher={Sage Publications Sage UK: London, England}
}

@inproceedings{he2016deep,
  title={Deep residual learning for image recognition},
  author={He, Kaiming and Zhang, Xiangyu and Ren, Shaoqing and Sun, Jian},
  booktitle={Proceedings of the IEEE conference on computer vision and pattern recognition},
  pages={770--778},
  year={2016}
}

@article{achiam2023gpt,
  title={Gpt-4 technical report},
  author={Achiam, Josh and Adler, Steven and Agarwal, Sandhini and Ahmad, Lama and Akkaya, Ilge and Aleman, Florencia Leoni and Almeida, Diogo and Altenschmidt, Janko and Altman, Sam and Anadkat, Shyamal and others},
  journal={arXiv preprint arXiv:2303.08774},
  year={2023}
}

@article{tian2024drivevlm,
  title={Drivevlm: The convergence of autonomous driving and large vision-language models},
  author={Tian, Xiaoyu and Gu, Junru and Li, Bailin and Liu, Yicheng and Wang, Yang and Zhao, Zhiyong and Zhan, Kun and Jia, Peng and Lang, Xianpeng and Zhao, Hang},
  journal={arXiv preprint arXiv:2402.12289},
  year={2024}
}

@inproceedings{chen2024internvl,
  title={Internvl: Scaling up vision foundation models and aligning for generic visual-linguistic tasks},
  author={Chen, Zhe and Wu, Jiannan and Wang, Wenhai and Su, Weijie and Chen, Guo and Xing, Sen and Zhong, Muyan and Zhang, Qinglong and Zhu, Xizhou and Lu, Lewei and others},
  booktitle={Proceedings of the IEEE/CVF conference on computer vision and pattern recognition},
  pages={24185--24198},
  year={2024}
}

@misc{li2025benchmarkingassessingsafetyrobustness,
      title={Towards Benchmarking and Assessing the Safety and Robustness of Autonomous Driving on Safety-critical Scenarios}, 
      author={Jingzheng Li and Xianglong Liu and Shikui Wei and Zhijun Chen and Bing Li and Qing Guo and Xianqi Yang and Yanjun Pu and Jiakai Wang},
      year={2025},
      eprint={2503.23708},
      archivePrefix={arXiv},
      primaryClass={cs.RO},
      url={https://arxiv.org/abs/2503.23708}, 
}

@misc{guo2024drivemllmbenchmarkspatialunderstanding,
      title={DriveMLLM: A Benchmark for Spatial Understanding with Multimodal Large Language Models in Autonomous Driving}, 
      author={Xianda Guo and Ruijun Zhang and Yiqun Duan and Yuhang He and Chenming Zhang and Shuai Liu and Long Chen},
      year={2024},
      eprint={2411.13112},
      archivePrefix={arXiv},
      primaryClass={cs.CV},
      url={https://arxiv.org/abs/2411.13112}, 
}

@article{schwarting_planning_2018,
	title = {Planning and {Decision}-{Making} for {Autonomous} {Vehicles}},
	volume = {1},
	issn = {2573-5144, 2573-5144},
	url = {https://www.annualreviews.org/doi/10.1146/annurev-control-060117-105157},
	doi = {10.1146/annurev-control-060117-105157},
	language = {en},
	number = {1},
	urldate = {2025-10-02},
	journal = {Annual Review of Control, Robotics, and Autonomous Systems},
	author = {Schwarting, Wilko and Alonso-Mora, Javier and Rus, Daniela},
	month = may,
	year = {2018},
	pages = {187--210},
}

@article{badue_self-driving_2021,
	title = {Self-driving cars: {A} survey},
	volume = {165},
	issn = {09574174},
	shorttitle = {Self-driving cars},
	url = {https://linkinghub.elsevier.com/retrieve/pii/S095741742030628X},
	doi = {10.1016/j.eswa.2020.113816},
	language = {en},
	urldate = {2025-10-02},
	journal = {Expert Systems with Applications},
	author = {Badue, Claudine and Guidolini, Rânik and Carneiro, Raphael Vivacqua and Azevedo, Pedro and Cardoso, Vinicius B. and Forechi, Avelino and Jesus, Luan and Berriel, Rodrigo and Paixão, Thiago M. and Mutz, Filipe and De Paula Veronese, Lucas and Oliveira-Santos, Thiago and De Souza, Alberto F.},
	month = mar,
	year = {2021},
	pages = {113816},
}

@article{paden_survey_2016,
	title = {A {Survey} of {Motion} {Planning} and {Control} {Techniques} for {Self}-{Driving} {Urban} {Vehicles}},
	volume = {1},
	issn = {2379-8904},
	url = {https://ieeexplore.ieee.org/document/7490340/},
	doi = {10.1109/TIV.2016.2578706},
	number = {1},
	urldate = {2025-10-02},
	journal = {IEEE Transactions on Intelligent Vehicles},
	author = {Paden, Brian and Čáp, Michal and Yong, Sze Zheng and Yershov, Dmitry and Frazzoli, Emilio},
	month = mar,
	year = {2016},
	pages = {33--55},
}

@misc{chen_end--end_2024,
	title = {End-to-end {Autonomous} {Driving}: {Challenges} and {Frontiers}},
	shorttitle = {End-to-end {Autonomous} {Driving}},
	url = {http://arxiv.org/abs/2306.16927},
	doi = {10.48550/arXiv.2306.16927},
	language = {en},
	urldate = {2025-10-02},
	publisher = {arXiv},
	author = {Chen, Li and Wu, Penghao and Chitta, Kashyap and Jaeger, Bernhard and Geiger, Andreas and Li, Hongyang},
	month = aug,
	year = {2024},
	note = {arXiv:2306.16927 [cs]},
}

@misc{codevilla_end--end_2018,
	title = {End-to-end {Driving} via {Conditional} {Imitation} {Learning}},
	url = {http://arxiv.org/abs/1710.02410},
	doi = {10.48550/arXiv.1710.02410},
	urldate = {2025-10-02},
	publisher = {arXiv},
	author = {Codevilla, Felipe and Müller, Matthias and López, Antonio and Koltun, Vladlen and Dosovitskiy, Alexey},
	month = mar,
	year = {2018},
	note = {arXiv:1710.02410 [cs]},
}

@misc{chib_recent_2023,
	title = {Recent {Advancements} in {End}-to-{End} {Autonomous} {Driving} using {Deep} {Learning}: {A} {Survey}},
	shorttitle = {Recent {Advancements} in {End}-to-{End} {Autonomous} {Driving} using {Deep} {Learning}},
	url = {http://arxiv.org/abs/2307.04370},
	doi = {10.48550/arXiv.2307.04370},
	urldate = {2025-10-02},
	publisher = {arXiv},
	author = {Chib, Pranav Singh and Singh, Pravendra},
	month = sep,
	year = {2023},
	note = {arXiv:2307.04370 [cs]},
}

@misc{zhao2026surveylargelanguagemodels,
      title={A Survey of Large Language Models}, 
      author={Wayne Xin Zhao and Kun Zhou and Junyi Li and Tianyi Tang and Xiaolei Wang and Yupeng Hou and Yingqian Min and Beichen Zhang and Junjie Zhang and Zican Dong and Yifan Du and Chen Yang and Yushuo Chen and Zhipeng Chen and Jinhao Jiang and Ruiyang Ren and Yifan Li and Xinyu Tang and Zikang Liu and Peiyu Liu and Jian-Yun Nie and Ji-Rong Wen},
      year={2026},
      eprint={2303.18223},
      archivePrefix={arXiv},
      primaryClass={cs.CL},
      url={https://arxiv.org/abs/2303.18223}, 
}

@misc{deepseek-ai_deepseek-v3_2025,
	title = {{DeepSeek}-{V3} {Technical} {Report}},
	url = {http://arxiv.org/abs/2412.19437},
	doi = {10.48550/arXiv.2412.19437},
	urldate = {2025-10-02},
	publisher = {arXiv},
	author = {DeepSeek-AI and Liu, Aixin and Feng, Bei and Xue, Bing and Wang, Bingxuan and Wu, Bochao and Lu, Chengda and Zhao, Chenggang and Deng, Chengqi and Zhang, Chenyu and Ruan, Chong and Dai, Damai and Guo, Daya and Yang, Dejian and Chen, Deli and Ji, Dongjie and Li, Erhang and Lin, Fangyun and Dai, Fucong and Luo, Fuli and Hao, Guangbo and Chen, Guanting and Li, Guowei and Zhang, H. and Bao, Han and Xu, Hanwei and Wang, Haocheng and Zhang, Haowei and Ding, Honghui and Xin, Huajian and Gao, Huazuo and Li, Hui and Qu, Hui and Cai, J. L. and Liang, Jian and Guo, Jianzhong and Ni, Jiaqi and Li, Jiashi and Wang, Jiawei and Chen, Jin and Chen, Jingchang and Yuan, Jingyang and Qiu, Junjie and Li, Junlong and Song, Junxiao and Dong, Kai and Hu, Kai and Gao, Kaige and Guan, Kang and Huang, Kexin and Yu, Kuai and Wang, Lean and Zhang, Lecong and Xu, Lei and Xia, Leyi and Zhao, Liang and Wang, Litong and Zhang, Liyue and Li, Meng and Wang, Miaojun and Zhang, Mingchuan and Zhang, Minghua and Tang, Minghui and Li, Mingming and Tian, Ning and Huang, Panpan and Wang, Peiyi and Zhang, Peng and Wang, Qiancheng and Zhu, Qihao and Chen, Qinyu and Du, Qiushi and Chen, R. J. and Jin, R. L. and Ge, Ruiqi and Zhang, Ruisong and Pan, Ruizhe and Wang, Runji and Xu, Runxin and Zhang, Ruoyu and Chen, Ruyi and Li, S. S. and Lu, Shanghao and Zhou, Shangyan and Chen, Shanhuang and Wu, Shaoqing and Ye, Shengfeng and Ye, Shengfeng and Ma, Shirong and Wang, Shiyu and Zhou, Shuang and Yu, Shuiping and Zhou, Shunfeng and Pan, Shuting and Wang, T. and Yun, Tao and Pei, Tian and Sun, Tianyu and Xiao, W. L. and Zeng, Wangding and Zhao, Wanjia and An, Wei and Liu, Wen and Liang, Wenfeng and Gao, Wenjun and Yu, Wenqin and Zhang, Wentao and Li, X. Q. and Jin, Xiangyue and Wang, Xianzu and Bi, Xiao and Liu, Xiaodong and Wang, Xiaohan and Shen, Xiaojin and Chen, Xiaokang and Zhang, Xiaokang and Chen, Xiaosha and Nie, Xiaotao and Sun, Xiaowen and Wang, Xiaoxiang and Cheng, Xin and Liu, Xin and Xie, Xin and Liu, Xingchao and Yu, Xingkai and Song, Xinnan and Shan, Xinxia and Zhou, Xinyi and Yang, Xinyu and Li, Xinyuan and Su, Xuecheng and Lin, Xuheng and Li, Y. K. and Wang, Y. Q. and Wei, Y. X. and Zhu, Y. X. and Zhang, Yang and Xu, Yanhong and Xu, Yanhong and Huang, Yanping and Li, Yao and Zhao, Yao and Sun, Yaofeng and Li, Yaohui and Wang, Yaohui and Yu, Yi and Zheng, Yi and Zhang, Yichao and Shi, Yifan and Xiong, Yiliang and He, Ying and Tang, Ying and Piao, Yishi and Wang, Yisong and Tan, Yixuan and Ma, Yiyang and Liu, Yiyuan and Guo, Yongqiang and Wu, Yu and Ou, Yuan and Zhu, Yuchen and Wang, Yuduan and Gong, Yue and Zou, Yuheng and He, Yujia and Zha, Yukun and Xiong, Yunfan and Ma, Yunxian and Yan, Yuting and Luo, Yuxiang and You, Yuxiang and Liu, Yuxuan and Zhou, Yuyang and Wu, Z. F. and Ren, Z. Z. and Ren, Zehui and Sha, Zhangli and Fu, Zhe and Xu, Zhean and Huang, Zhen and Zhang, Zhen and Xie, Zhenda and Zhang, Zhengyan and Hao, Zhewen and Gou, Zhibin and Ma, Zhicheng and Yan, Zhigang and Shao, Zhihong and Xu, Zhipeng and Wu, Zhiyu and Zhang, Zhongyu and Li, Zhuoshu and Gu, Zihui and Zhu, Zijia and Liu, Zijun and Li, Zilin and Xie, Ziwei and Song, Ziyang and Gao, Ziyi and Pan, Zizheng},
	month = feb,
	year = {2025},
	note = {arXiv:2412.19437 [cs]},
}

@misc{touvron_llama_2023,
	title = {{LLaMA}: {Open} and {Efficient} {Foundation} {Language} {Models}},
	shorttitle = {{LLaMA}},
	url = {http://arxiv.org/abs/2302.13971},
	doi = {10.48550/arXiv.2302.13971},
	urldate = {2025-10-02},
	publisher = {arXiv},
	author = {Touvron, Hugo and Lavril, Thibaut and Izacard, Gautier and Martinet, Xavier and Lachaux, Marie-Anne and Lacroix, Timothée and Rozière, Baptiste and Goyal, Naman and Hambro, Eric and Azhar, Faisal and Rodriguez, Aurelien and Joulin, Armand and Grave, Edouard and Lample, Guillaume},
	month = feb,
	year = {2023},
	note = {arXiv:2302.13971 [cs]},
}

@misc{openai_gpt-4_2024,
	title = {{GPT}-4 {Technical} {Report}},
	url = {http://arxiv.org/abs/2303.08774},
	doi = {10.48550/arXiv.2303.08774},
	urldate = {2025-10-02},
	publisher = {arXiv},
	author = {OpenAI and Achiam, Josh and Adler, Steven and Agarwal, Sandhini and Ahmad, Lama and Akkaya, Ilge and Aleman, Florencia Leoni and Almeida, Diogo and Altenschmidt, Janko and Altman, Sam and Anadkat, Shyamal and Avila, Red and Babuschkin, Igor and Balaji, Suchir and Balcom, Valerie and Baltescu, Paul and Bao, Haiming and Bavarian, Mohammad and Belgum, Jeff and Bello, Irwan and Berdine, Jake and Bernadett-Shapiro, Gabriel and Berner, Christopher and Bogdonoff, Lenny and Boiko, Oleg and Boyd, Madelaine and Brakman, Anna-Luisa and Brockman, Greg and Brooks, Tim and Brundage, Miles and Button, Kevin and Cai, Trevor and Campbell, Rosie and Cann, Andrew and Carey, Brittany and Carlson, Chelsea and Carmichael, Rory and Chan, Brooke and Chang, Che and Chantzis, Fotis and Chen, Derek and Chen, Sully and Chen, Ruby and Chen, Jason and Chen, Mark and Chess, Ben and Cho, Chester and Chu, Casey and Chung, Hyung Won and Cummings, Dave and Currier, Jeremiah and Dai, Yunxing and Decareaux, Cory and Degry, Thomas and Deutsch, Noah and Deville, Damien and Dhar, Arka and Dohan, David and Dowling, Steve and Dunning, Sheila and Ecoffet, Adrien and Eleti, Atty and Eloundou, Tyna and Farhi, David and Fedus, Liam and Felix, Niko and Fishman, Simón Posada and Forte, Juston and Fulford, Isabella and Gao, Leo and Georges, Elie and Gibson, Christian and Goel, Vik and Gogineni, Tarun and Goh, Gabriel and Gontijo-Lopes, Rapha and Gordon, Jonathan and Grafstein, Morgan and Gray, Scott and Greene, Ryan and Gross, Joshua and Gu, Shixiang Shane and Guo, Yufei and Hallacy, Chris and Han, Jesse and Harris, Jeff and He, Yuchen and Heaton, Mike and Heidecke, Johannes and Hesse, Chris and Hickey, Alan and Hickey, Wade and Hoeschele, Peter and Houghton, Brandon and Hsu, Kenny and Hu, Shengli and Hu, Xin and Huizinga, Joost and Jain, Shantanu and Jain, Shawn and Jang, Joanne and Jiang, Angela and Jiang, Roger and Jin, Haozhun and Jin, Denny and Jomoto, Shino and Jonn, Billie and Jun, Heewoo and Kaftan, Tomer and Kaiser, Łukasz and Kamali, Ali and Kanitscheider, Ingmar and Keskar, Nitish Shirish and Khan, Tabarak and Kilpatrick, Logan and Kim, Jong Wook and Kim, Christina and Kim, Yongjik and Kirchner, Jan Hendrik and Kiros, Jamie and Knight, Matt and Kokotajlo, Daniel and Kondraciuk, Łukasz and Kondrich, Andrew and Konstantinidis, Aris and Kosic, Kyle and Krueger, Gretchen and Kuo, Vishal and Lampe, Michael and Lan, Ikai and Lee, Teddy and Leike, Jan and Leung, Jade and Levy, Daniel and Li, Chak Ming and Lim, Rachel and Lin, Molly and Lin, Stephanie and Litwin, Mateusz and Lopez, Theresa and Lowe, Ryan and Lue, Patricia and Makanju, Anna and Malfacini, Kim and Manning, Sam and Markov, Todor and Markovski, Yaniv and Martin, Bianca and Mayer, Katie and Mayne, Andrew and McGrew, Bob and McKinney, Scott Mayer and McLeavey, Christine and McMillan, Paul and McNeil, Jake and Medina, David and Mehta, Aalok and Menick, Jacob and Metz, Luke and Mishchenko, Andrey and Mishkin, Pamela and Monaco, Vinnie and Morikawa, Evan and Mossing, Daniel and Mu, Tong and Murati, Mira and Murk, Oleg and Mély, David and Nair, Ashvin and Nakano, Reiichiro and Nayak, Rajeev and Neelakantan, Arvind and Ngo, Richard and Noh, Hyeonwoo and Ouyang, Long and O'Keefe, Cullen and Pachocki, Jakub and Paino, Alex and Palermo, Joe and Pantuliano, Ashley and Parascandolo, Giambattista and Parish, Joel and Parparita, Emy and Passos, Alex and Pavlov, Mikhail and Peng, Andrew and Perelman, Adam and Peres, Filipe de Avila Belbute and Petrov, Michael and Pinto, Henrique Ponde de Oliveira and Michael and Pokorny and Pokrass, Michelle and Pong, Vitchyr H. and Powell, Tolly and Power, Alethea and Power, Boris and Proehl, Elizabeth and Puri, Raul and Radford, Alec and Rae, Jack and Ramesh, Aditya and Raymond, Cameron and Real, Francis and Rimbach, Kendra and Ross, Carl and Rotsted, Bob and Roussez, Henri and Ryder, Nick and Saltarelli, Mario and Sanders, Ted and Santurkar, Shibani and Sastry, Girish and Schmidt, Heather and Schnurr, David and Schulman, John and Selsam, Daniel and Sheppard, Kyla and Sherbakov, Toki and Shieh, Jessica and Shoker, Sarah and Shyam, Pranav and Sidor, Szymon and Sigler, Eric and Simens, Maddie and Sitkin, Jordan and Slama, Katarina and Sohl, Ian and Sokolowsky, Benjamin and Song, Yang and Staudacher, Natalie and Such, Felipe Petroski and Summers, Natalie and Sutskever, Ilya and Tang, Jie and Tezak, Nikolas and Thompson, Madeleine B. and Tillet, Phil and Tootoonchian, Amin and Tseng, Elizabeth and Tuggle, Preston and Turley, Nick and Tworek, Jerry and Uribe, Juan Felipe Cerón and Vallone, Andrea and Vijayvergiya, Arun and Voss, Chelsea and Wainwright, Carroll and Wang, Justin Jay and Wang, Alvin and Wang, Ben and Ward, Jonathan and Wei, Jason and Weinmann, C. J. and Welihinda, Akila and Welinder, Peter and Weng, Jiayi and Weng, Lilian and Wiethoff, Matt and Willner, Dave and Winter, Clemens and Wolrich, Samuel and Wong, Hannah and Workman, Lauren and Wu, Sherwin and Wu, Jeff and Wu, Michael and Xiao, Kai and Xu, Tao and Yoo, Sarah and Yu, Kevin and Yuan, Qiming and Zaremba, Wojciech and Zellers, Rowan and Zhang, Chong and Zhang, Marvin and Zhao, Shengjia and Zheng, Tianhao and Zhuang, Juntang and Zhuk, William and Zoph, Barret},
	month = mar,
	year = {2024},
	note = {arXiv:2303.08774 [cs]
version: 6},
}

@inproceedings{shen_aligning_2024,
	address = {Seattle, WA, USA},
	title = {Aligning and {Prompting} {Everything} {All} at {Once} for {Universal} {Visual} {Perception}},
	copyright = {https://doi.org/10.15223/policy-029},
	isbn = {979-8-3503-5300-6},
	url = {https://ieeexplore.ieee.org/document/10658173/},
	doi = {10.1109/CVPR52733.2024.01253},
	language = {en},
	urldate = {2025-10-02},
	booktitle = {2024 {IEEE}/{CVF} {Conference} on {Computer} {Vision} and {Pattern} {Recognition} ({CVPR})},
	publisher = {IEEE},
	author = {Shen, Yunhang and Fu, Chaoyou and Chen, Peixian and Zhang, Mengdan and Li, Ke and Sun, Xing and Wu, Yunsheng and Lin, Shaohui and Ji, Rongrong},
	month = jun,
	year = {2024},
	pages = {13193--13203},
}

@article{zhang_vision-language_2024,
	title = {Vision-{Language} {Models} for {Vision} {Tasks}: {A} {Survey}},
	volume = {46},
	issn = {1939-3539},
	shorttitle = {Vision-{Language} {Models} for {Vision} {Tasks}},
	url = {https://ieeexplore.ieee.org/document/10445007/},
	doi = {10.1109/TPAMI.2024.3369699},
	number = {8},
	urldate = {2025-10-02},
	journal = {IEEE Transactions on Pattern Analysis and Machine Intelligence},
	author = {Zhang, Jingyi and Huang, Jiaxing and Jin, Sheng and Lu, Shijian},
	month = aug,
	year = {2024},
	pages = {5625--5644},
}

@misc{lu_internvl-x_2025,
	title = {{InternVL}-{X}: {Advancing} and {Accelerating} {InternVL} {Series} with {Efficient} {Visual} {Token} {Compression}},
	shorttitle = {{InternVL}-{X}},
	url = {http://arxiv.org/abs/2503.21307},
	doi = {10.48550/arXiv.2503.21307},
	urldate = {2025-10-02},
	publisher = {arXiv},
	author = {Lu, Dongchen and Sun, Yuyao and Zhang, Zilu and Huang, Leping and Zeng, Jianliang and Shu, Mao and Cao, Huo},
	month = mar,
	year = {2025},
	note = {arXiv:2503.21307 [cs]},
}

@misc{you_v2x-vlm_2025,
	title = {{V2X}-{VLM}: {End}-to-{End} {V2X} {Cooperative} {Autonomous} {Driving} {Through} {Large} {Vision}-{Language} {Models}},
	shorttitle = {{V2X}-{VLM}},
	url = {http://arxiv.org/abs/2408.09251},
	doi = {10.48550/arXiv.2408.09251},
	urldate = {2025-10-02},
	publisher = {arXiv},
	author = {You, Junwei and Shi, Haotian and Jiang, Zhuoyu and Huang, Zilin and Gan, Rui and Wu, Keshu and Cheng, Xi and Li, Xiaopeng and Ran, Bin},
	month = jun,
	year = {2025},
	note = {arXiv:2408.09251 [cs]},
}

@article{zhou_vision_2024,
	title = {Vision {Language} {Models} in {Autonomous} {Driving}: {A} {Survey} and {Outlook}},
	issn = {2379-8904},
	shorttitle = {Vision {Language} {Models} in {Autonomous} {Driving}},
	url = {https://ieeexplore.ieee.org/document/10531702/},
	doi = {10.1109/TIV.2024.3402136},
	urldate = {2025-10-02},
	journal = {IEEE Transactions on Intelligent Vehicles},
	author = {Zhou, Xingcheng and Liu, Mingyu and Yurtsever, Ekim and Zagar, Bare Luka and Zimmer, Walter and Cao, Hu and Knoll, Alois C.},
	year = {2024},
	pages = {1--20},
}

@misc{tian_drivevlm_2024,
	title = {{DriveVLM}: {The} {Convergence} of {Autonomous} {Driving} and {Large} {Vision}-{Language} {Models}},
	shorttitle = {{DriveVLM}},
	url = {http://arxiv.org/abs/2402.12289},
	doi = {10.48550/arXiv.2402.12289},
	urldate = {2025-10-02},
	publisher = {arXiv},
	author = {Tian, Xiaoyu and Gu, Junru and Li, Bailin and Liu, Yicheng and Wang, Yang and Zhao, Zhiyong and Zhan, Kun and Jia, Peng and Lang, Xianpeng and Zhao, Hang},
	month = jun,
	year = {2024},
	note = {arXiv:2402.12289 [cs]},
}

@inproceedings{fu_drivegenvlm_2024,
	title = {{DriveGenVLM}: {Real}-world {Video} {Generation} for {Vision} {Language} {Model} based {Autonomous} {Driving}},
	shorttitle = {{DriveGenVLM}},
	url = {https://ieeexplore.ieee.org/document/10786438/},
	doi = {10.1109/IAVVC63304.2024.10786438},
	urldate = {2025-10-02},
	booktitle = {2024 {IEEE} {International} {Automated} {Vehicle} {Validation} {Conference} ({IAVVC})},
	author = {Fu, Yongjie and Jain, Anmol and Chen, Xu and Mo, Zhaobin and Di, Xuan},
	month = oct,
	year = {2024},
	pages = {1--6},
}

@misc{gopalkrishnan_multi-frame_2024,
	title = {Multi-{Frame}, {Lightweight} \& {Efficient} {Vision}-{Language} {Models} for {Question} {Answering} in {Autonomous} {Driving}},
	url = {http://arxiv.org/abs/2403.19838},
	doi = {10.48550/arXiv.2403.19838},
	urldate = {2025-10-02},
	publisher = {arXiv},
	author = {Gopalkrishnan, Akshay and Greer, Ross and Trivedi, Mohan},
	month = may,
	year = {2024},
	note = {arXiv:2403.19838 [cs]},
}

@inproceedings{li_blip-2_2023,
	title = {{BLIP}-2: {Bootstrapping} {Language}-{Image} {Pre}-training with {Frozen} {Image} {Encoders} and {Large} {Language} {Models}},
	shorttitle = {{BLIP}-2},
	url = {https://proceedings.mlr.press/v202/li23q.html},
	language = {en},
	urldate = {2025-10-02},
	booktitle = {Proceedings of the 40th {International} {Conference} on {Machine} {Learning}},
	publisher = {PMLR},
	author = {Li, Junnan and Li, Dongxu and Savarese, Silvio and Hoi, Steven},
	month = jul,
	year = {2023},
	note = {ISSN: 2640-3498},
	pages = {19730--19742},
}

@misc{wang_qwen2-vl_2024,
	title = {Qwen2-{VL}: {Enhancing} {Vision}-{Language} {Model}'s {Perception} of the {World} at {Any} {Resolution}},
	shorttitle = {Qwen2-{VL}},
	url = {http://arxiv.org/abs/2409.12191},
	doi = {10.48550/arXiv.2409.12191},
	urldate = {2025-10-02},
	publisher = {arXiv},
	author = {Wang, Peng and Bai, Shuai and Tan, Sinan and Wang, Shijie and Fan, Zhihao and Bai, Jinze and Chen, Keqin and Liu, Xuejing and Wang, Jialin and Ge, Wenbin and Fan, Yang and Dang, Kai and Du, Mengfei and Ren, Xuancheng and Men, Rui and Liu, Dayiheng and Zhou, Chang and Zhou, Jingren and Lin, Junyang},
	month = oct,
	year = {2024},
	note = {arXiv:2409.12191 [cs]},
}

@inproceedings{sima_drivelm_2025,
	address = {Cham},
	title = {{DriveLM}: {Driving} with {Graph} {Visual} {Question} {Answering}},
	isbn = {978-3-031-72943-0},
	booktitle = {Computer {Vision} – {ECCV} 2024},
	publisher = {Springer Nature Switzerland},
	author = {Sima, Chonghao and Renz, Katrin and Chitta, Kashyap and Chen, Li and Zhang, Hanxue and Xie, Chengen and Beißwenger, Jens and Luo, Ping and Geiger, Andreas and Li, Hongyang},
	editor = {Leonardis, Aleš and Ricci, Elisa and Roth, Stefan and Russakovsky, Olga and Sattler, Torsten and Varol, Gül},
	year = {2025},
	pages = {256--274},
}

@article{xu2024drivegpt4,
  title={Drivegpt4: Interpretable end-to-end autonomous driving via large language model},
  author={Xu, Zhenhua and Zhang, Yujia and Xie, Enze and Zhao, Zhen and Guo, Yong and Wong, Kwan-Yee K and Li, Zhenguo and Zhao, Hengshuang},
  journal={IEEE Robotics and Automation Letters},
  year={2024},
  publisher={IEEE}
}

@inproceedings{guan_hallusionbench_2024,
	address = {Seattle, WA, USA},
	title = {Hallusionbench: {An} {Advanced} {Diagnostic} {Suite} for {Entangled} {Language} {Hallucination} and {Visual} {Illusion} in {Large} {Vision}-{Language} {Models}},
	copyright = {https://doi.org/10.15223/policy-029},
	isbn = {979-8-3503-5300-6},
	shorttitle = {Hallusionbench},
	url = {https://ieeexplore.ieee.org/document/10657594/},
	doi = {10.1109/CVPR52733.2024.01363},
	language = {en},
	urldate = {2025-10-02},
	booktitle = {2024 {IEEE}/{CVF} {Conference} on {Computer} {Vision} and {Pattern} {Recognition} ({CVPR})},
	publisher = {IEEE},
	author = {Guan, Tianrui and Liu, Fuxiao and Wu, Xiyang and Xian, Ruiqi and Li, Zongxia and Liu, Xiaoyu and Wang, Xijun and Chen, Lichang and Huang, Furong and Yacoob, Yaser and Manocha, Dinesh and Zhou, Tianyi},
	month = jun,
	year = {2024},
	pages = {14375--14385},
}

@misc{yu_mm-vet_2024,
	title = {{MM}-{Vet}: {Evaluating} {Large} {Multimodal} {Models} for {Integrated} {Capabilities}},
	shorttitle = {{MM}-{Vet}},
	url = {http://arxiv.org/abs/2308.02490},
	doi = {10.48550/arXiv.2308.02490},
	urldate = {2025-10-02},
	publisher = {arXiv},
	author = {Yu, Weihao and Yang, Zhengyuan and Li, Linjie and Wang, Jianfeng and Lin, Kevin and Liu, Zicheng and Wang, Xinchao and Wang, Lijuan},
	month = dec,
	year = {2024},
	note = {arXiv:2308.02490 [cs]},
}

@misc{bai_hallucination_2025,
	title = {Hallucination of {Multimodal} {Large} {Language} {Models}: {A} {Survey}},
	shorttitle = {Hallucination of {Multimodal} {Large} {Language} {Models}},
	url = {http://arxiv.org/abs/2404.18930},
	doi = {10.48550/arXiv.2404.18930},
	urldate = {2025-10-02},
	publisher = {arXiv},
	author = {Bai, Zechen and Wang, Pichao and Xiao, Tianjun and He, Tong and Han, Zongbo and Zhang, Zheng and Shou, Mike Zheng},
	month = apr,
	year = {2025},
	note = {arXiv:2404.18930 [cs]},
}

@inproceedings{wang_mmlu-pro_2024,
	title = {{MMLU}-{Pro}: {A} {More} {Robust} and {Challenging} {Multi}-{Task} {Language} {Understanding} {Benchmark}},
	volume = {37},
	url = {https://proceedings.neurips.cc/paper_files/paper/2024/file/ad236edc564f3e3156e1b2feafb99a24-Paper-Datasets_and_Benchmarks_Track.pdf},
	booktitle = {Advances in {Neural} {Information} {Processing} {Systems}},
	publisher = {Curran Associates, Inc.},
	author = {Wang, Yubo and Ma, Xueguang and Zhang, Ge and Ni, Yuansheng and Chandra, Abhranil and Guo, Shiguang and Ren, Weiming and Arulraj, Aaran and He, Xuan and Jiang, Ziyan and Li, Tianle and Ku, Max and Wang, Kai and Zhuang, Alex and Fan, Rongqi and Yue, Xiang and Chen, Wenhu},
	editor = {Globerson, A. and Mackey, L. and Belgrave, D. and Fan, A. and Paquet, U. and Tomczak, J. and Zhang, C.},
	year = {2024},
	pages = {95266--95290},
}

@article{rein2023gpqa,
  title={Gpqa: A graduate-level google-proof q\&a benchmark},
  author={Rein, David and Hou, Betty Li and Stickland, Asa Cooper and Petty, Jackson and Pang, Richard Yuanzhe and Dirani, Julien and Michael, Julian and Bowman, Samuel R},
  journal={arXiv preprint arXiv:2311.12022},
  year={2023}
}

@misc{patel_aime_2024,
	title = {{AIME}: {AI} {System} {Optimization} via {Multiple} {LLM} {Evaluators}},
	shorttitle = {{AIME}},
	url = {http://arxiv.org/abs/2410.03131},
	doi = {10.48550/arXiv.2410.03131},
	urldate = {2025-10-02},
	publisher = {arXiv},
	author = {Patel, Bhrij and Chakraborty, Souradip and Suttle, Wesley A. and Wang, Mengdi and Bedi, Amrit Singh and Manocha, Dinesh},
	month = oct,
	year = {2024},
	note = {arXiv:2410.03131 [cs]},
}

@article{sakaguchi_winogrande_2021,
	title = {{WinoGrande}: an adversarial winograd schema challenge at scale},
	volume = {64},
	issn = {0001-0782, 1557-7317},
	shorttitle = {{WinoGrande}},
	url = {https://dl.acm.org/doi/10.1145/3474381},
	doi = {10.1145/3474381},
	language = {en},
	number = {9},
	urldate = {2025-10-02},
	journal = {Communications of the ACM},
	author = {Sakaguchi, Keisuke and Bras, Ronan Le and Bhagavatula, Chandra and Choi, Yejin},
	month = sep,
	year = {2021},
	pages = {99--106},
}

@misc{clark_think_2018,
	title = {Think you have {Solved} {Question} {Answering}? {Try} {ARC}, the {AI2} {Reasoning} {Challenge}},
	shorttitle = {Think you have {Solved} {Question} {Answering}?},
	url = {http://arxiv.org/abs/1803.05457},
	doi = {10.48550/arXiv.1803.05457},
	urldate = {2025-10-02},
	publisher = {arXiv},
	author = {Clark, Peter and Cowhey, Isaac and Etzioni, Oren and Khot, Tushar and Sabharwal, Ashish and Schoenick, Carissa and Tafjord, Oyvind},
	month = mar,
	year = {2018},
	note = {arXiv:1803.05457 [cs]},
}

@misc{chen_evaluating_2021,
	title = {Evaluating {Large} {Language} {Models} {Trained} on {Code}},
	url = {http://arxiv.org/abs/2107.03374},
	doi = {10.48550/arXiv.2107.03374},
	urldate = {2025-10-02},
	publisher = {arXiv},
	author = {Chen, Mark and Tworek, Jerry and Jun, Heewoo and Yuan, Qiming and Pinto, Henrique Ponde de Oliveira and Kaplan, Jared and Edwards, Harri and Burda, Yuri and Joseph, Nicholas and Brockman, Greg and Ray, Alex and Puri, Raul and Krueger, Gretchen and Petrov, Michael and Khlaaf, Heidy and Sastry, Girish and Mishkin, Pamela and Chan, Brooke and Gray, Scott and Ryder, Nick and Pavlov, Mikhail and Power, Alethea and Kaiser, Lukasz and Bavarian, Mohammad and Winter, Clemens and Tillet, Philippe and Such, Felipe Petroski and Cummings, Dave and Plappert, Matthias and Chantzis, Fotios and Barnes, Elizabeth and Herbert-Voss, Ariel and Guss, William Hebgen and Nichol, Alex and Paino, Alex and Tezak, Nikolas and Tang, Jie and Babuschkin, Igor and Balaji, Suchir and Jain, Shantanu and Saunders, William and Hesse, Christopher and Carr, Andrew N. and Leike, Jan and Achiam, Josh and Misra, Vedant and Morikawa, Evan and Radford, Alec and Knight, Matthew and Brundage, Miles and Murati, Mira and Mayer, Katie and Welinder, Peter and McGrew, Bob and Amodei, Dario and McCandlish, Sam and Sutskever, Ilya and Zaremba, Wojciech},
	month = jul,
	year = {2021},
	note = {arXiv:2107.03374 [cs]},
}

@misc{tang_mtvqa_2025,
	title = {{MTVQA}: {Benchmarking} {Multilingual} {Text}-{Centric} {Visual} {Question} {Answering}},
	shorttitle = {{MTVQA}},
	url = {http://arxiv.org/abs/2405.11985},
	doi = {10.48550/arXiv.2405.11985},
	urldate = {2025-10-02},
	publisher = {arXiv},
	author = {Tang, Jingqun and Liu, Qi and Ye, Yongjie and Lu, Jinghui and Wei, Shu and Lin, Chunhui and Li, Wanqing and Mahmood, Mohamad Fitri Faiz Bin and Feng, Hao and Zhao, Zhen and He, Yangfan and Lu, Kuan and Wang, Yanjie and Liu, Yuliang and Liu, Hao and Bai, Xiang and Huang, Can},
	month = jun,
	year = {2025},
	note = {arXiv:2405.11985 [cs]},
}

@inproceedings{yue_mmmu_2024,
	address = {Seattle, WA, USA},
	title = {{MMMU}: {A} {Massive} {Multi}-{Discipline} {Multimodal} {Understanding} and {Reasoning} {Benchmark} for {Expert} {AGI}},
	copyright = {https://doi.org/10.15223/policy-029},
	isbn = {979-8-3503-5300-6},
	shorttitle = {{MMMU}},
	url = {https://ieeexplore.ieee.org/document/10656299/},
	doi = {10.1109/CVPR52733.2024.00913},
	language = {en},
	urldate = {2025-10-02},
	booktitle = {2024 {IEEE}/{CVF} {Conference} on {Computer} {Vision} and {Pattern} {Recognition} ({CVPR})},
	publisher = {IEEE},
	author = {Yue, Xiang and Ni, Yuansheng and Zheng, Tianyu and Zhang, Kai and Liu, Ruoqi and Zhang, Ge and Stevens, Samuel and Jiang, Dongfu and Ren, Weiming and Sun, Yuxuan and Wei, Cong and Yu, Botao and Yuan, Ruibin and Sun, Renliang and Yin, Ming and Zheng, Boyuan and Yang, Zhenzhu and Liu, Yibo and Huang, Wenhao and Sun, Huan and Su, Yu and Chen, Wenhu},
	month = jun,
	year = {2024},
	pages = {9556--9567},
}

@article{liu_ocrbench_2024,
	title = {{OCRBench}: on the hidden mystery of {OCR} in large multimodal models},
	volume = {67},
	issn = {1869-1919},
	url = {https://doi.org/10.1007/s11432-024-4235-6},
	doi = {10.1007/s11432-024-4235-6},
	number = {12},
	journal = {Science China Information Sciences},
	author = {Liu, Yuliang and Li, Zhang and Huang, Mingxin and Yang, Biao and Yu, Wenwen and Li, Chunyuan and Yin, Xu-Cheng and Liu, Cheng-Lin and Jin, Lianwen and Bai, Xiang},
	month = dec,
	year = {2024},
	pages = {220102},
}

@misc{zhang_vcr_2025,
	title = {{VCR}: {A} {Task} for {Pixel}-{Level} {Complex} {Reasoning} in {Vision} {Language} {Models} via {Restoring} {Occluded} {Text}},
	shorttitle = {{VCR}},
	url = {http://arxiv.org/abs/2406.06462},
	doi = {10.48550/arXiv.2406.06462},
	urldate = {2025-10-02},
	publisher = {arXiv},
	author = {Zhang, Tianyu and Wang, Suyuchen and Li, Lu and Zhang, Ge and Taslakian, Perouz and Rajeswar, Sai and Fu, Jie and Liu, Bang and Bengio, Yoshua},
	month = apr,
	year = {2025},
	note = {arXiv:2406.06462 [cs]},
}

@misc{yue_mmmu-pro_2025,
	title = {{MMMU}-{Pro}: {A} {More} {Robust} {Multi}-discipline {Multimodal} {Understanding} {Benchmark}},
	shorttitle = {{MMMU}-{Pro}},
	url = {http://arxiv.org/abs/2409.02813},
	doi = {10.48550/arXiv.2409.02813},
	urldate = {2025-10-02},
	publisher = {arXiv},
	author = {Yue, Xiang and Zheng, Tianyu and Ni, Yuansheng and Wang, Yubo and Zhang, Kai and Tong, Shengbang and Sun, Yuxuan and Yu, Botao and Zhang, Ge and Sun, Huan and Su, Yu and Chen, Wenhu and Neubig, Graham},
	month = may,
	year = {2025},
	note = {arXiv:2409.02813 [cs]},
}

@misc{liu2023improved,
      title={Improved Baselines with Visual Instruction Tuning}, 
      author={Haotian Liu and Chunyuan Li and Yuheng Li and Yong Jae Lee},
      year={2023},
      eprint={2310.03744},
      archivePrefix={arXiv},
      primaryClass={cs.CV}
}

@misc{bai2025qwen25vltechnicalreport,
      title={Qwen2.5-VL Technical Report}, 
      author={Shuai Bai and Keqin Chen and Xuejing Liu and Jialin Wang and Wenbin Ge and Sibo Song and Kai Dang and Peng Wang and Shijie Wang and Jun Tang and Humen Zhong and Yuanzhi Zhu and Mingkun Yang and Zhaohai Li and Jianqiang Wan and Pengfei Wang and Wei Ding and Zheren Fu and Yiheng Xu and Jiabo Ye and Xi Zhang and Tianbao Xie and Zesen Cheng and Hang Zhang and Zhibo Yang and Haiyang Xu and Junyang Lin},
      year={2025},
      eprint={2502.13923},
      archivePrefix={arXiv},
      primaryClass={cs.CV},
      url={https://arxiv.org/abs/2502.13923}, 
}

@misc{gemmateam2025gemma3technicalreport,
      title={Gemma 3 Technical Report}, 
      author={Gemma Team and Aishwarya Kamath and Johan Ferret and Shreya Pathak and Nino Vieillard and Ramona Merhej and Sarah Perrin and Tatiana Matejovicova and Alexandre Ramé and Morgane Rivière and Louis Rouillard and Thomas Mesnard and Geoffrey Cideron and Jean-bastien Grill and Sabela Ramos and Edouard Yvinec and Michelle Casbon and Etienne Pot and Ivo Penchev and Gaël Liu and Francesco Visin and Kathleen Kenealy and Lucas Beyer and Xiaohai Zhai and Anton Tsitsulin and Robert Busa-Fekete and Alex Feng and Noveen Sachdeva and Benjamin Coleman and Yi Gao and Basil Mustafa and Iain Barr and Emilio Parisotto and David Tian and Matan Eyal and Colin Cherry and Jan-Thorsten Peter and Danila Sinopalnikov and Surya Bhupatiraju and Rishabh Agarwal and Mehran Kazemi and Dan Malkin and Ravin Kumar and David Vilar and Idan Brusilovsky and Jiaming Luo and Andreas Steiner and Abe Friesen and Abhanshu Sharma and Abheesht Sharma and Adi Mayrav Gilady and Adrian Goedeckemeyer and Alaa Saade and Alex Feng and Alexander Kolesnikov and Alexei Bendebury and Alvin Abdagic and Amit Vadi and András György and André Susano Pinto and Anil Das and Ankur Bapna and Antoine Miech and Antoine Yang and Antonia Paterson and Ashish Shenoy and Ayan Chakrabarti and Bilal Piot and Bo Wu and Bobak Shahriari and Bryce Petrini and Charlie Chen and Charline Le Lan and Christopher A. Choquette-Choo and CJ Carey and Cormac Brick and Daniel Deutsch and Danielle Eisenbud and Dee Cattle and Derek Cheng and Dimitris Paparas and Divyashree Shivakumar Sreepathihalli and Doug Reid and Dustin Tran and Dustin Zelle and Eric Noland and Erwin Huizenga and Eugene Kharitonov and Frederick Liu and Gagik Amirkhanyan and Glenn Cameron and Hadi Hashemi and Hanna Klimczak-Plucińska and Harman Singh and Harsh Mehta and Harshal Tushar Lehri and Hussein Hazimeh and Ian Ballantyne and Idan Szpektor and Ivan Nardini and Jean Pouget-Abadie and Jetha Chan and Joe Stanton and John Wieting and Jonathan Lai and Jordi Orbay and Joseph Fernandez and Josh Newlan and Ju-yeong Ji and Jyotinder Singh and Kat Black and Kathy Yu and Kevin Hui and Kiran Vodrahalli and Klaus Greff and Linhai Qiu and Marcella Valentine and Marina Coelho and Marvin Ritter and Matt Hoffman and Matthew Watson and Mayank Chaturvedi and Michael Moynihan and Min Ma and Nabila Babar and Natasha Noy and Nathan Byrd and Nick Roy and Nikola Momchev and Nilay Chauhan and Noveen Sachdeva and Oskar Bunyan and Pankil Botarda and Paul Caron and Paul Kishan Rubenstein and Phil Culliton and Philipp Schmid and Pier Giuseppe Sessa and Pingmei Xu and Piotr Stanczyk and Pouya Tafti and Rakesh Shivanna and Renjie Wu and Renke Pan and Reza Rokni and Rob Willoughby and Rohith Vallu and Ryan Mullins and Sammy Jerome and Sara Smoot and Sertan Girgin and Shariq Iqbal and Shashir Reddy and Shruti Sheth and Siim Põder and Sijal Bhatnagar and Sindhu Raghuram Panyam and Sivan Eiger and Susan Zhang and Tianqi Liu and Trevor Yacovone and Tyler Liechty and Uday Kalra and Utku Evci and Vedant Misra and Vincent Roseberry and Vlad Feinberg and Vlad Kolesnikov and Woohyun Han and Woosuk Kwon and Xi Chen and Yinlam Chow and Yuvein Zhu and Zichuan Wei and Zoltan Egyed and Victor Cotruta and Minh Giang and Phoebe Kirk and Anand Rao and Kat Black and Nabila Babar and Jessica Lo and Erica Moreira and Luiz Gustavo Martins and Omar Sanseviero and Lucas Gonzalez and Zach Gleicher and Tris Warkentin and Vahab Mirrokni and Evan Senter and Eli Collins and Joelle Barral and Zoubin Ghahramani and Raia Hadsell and Yossi Matias and D. Sculley and Slav Petrov and Noah Fiedel and Noam Shazeer and Oriol Vinyals and Jeff Dean and Demis Hassabis and Koray Kavukcuoglu and Clement Farabet and Elena Buchatskaya and Jean-Baptiste Alayrac and Rohan Anil and Dmitry and Lepikhin and Sebastian Borgeaud and Olivier Bachem and Armand Joulin and Alek Andreev and Cassidy Hardin and Robert Dadashi and Léonard Hussenot},
      year={2025},
      eprint={2503.19786},
      archivePrefix={arXiv},
      primaryClass={cs.CL},
      url={https://arxiv.org/abs/2503.19786}, 
}

@ARTICLE{2024arXiv240721783G,
       author = {{Grattafiori}, Aaron and {Dubey}, Abhimanyu and {Jauhri}, Abhinav and {Pandey}, Abhinav and {Kadian}, Abhishek and {Al-Dahle}, Ahmad and {Letman}, Aiesha and {Mathur}, Akhil and {Schelten}, Alan and {Vaughan}, Alex and {Yang}, Amy and {Fan}, Angela and {Goyal}, Anirudh and {Hartshorn}, Anthony and {Yang}, Aobo and {Mitra}, Archi and {Sravankumar}, Archie and {Korenev}, Artem and {Hinsvark}, Arthur and {Rao}, Arun and {Zhang}, Aston and {Rodriguez}, Aurelien and {Gregerson}, Austen and {Spataru}, Ava and {Roziere}, Baptiste and {Biron}, Bethany and {Tang}, Binh and {Chern}, Bobbie and {Caucheteux}, Charlotte and {Nayak}, Chaya and {Bi}, Chloe and {Marra}, Chris and {McConnell}, Chris and {Keller}, Christian and {Touret}, Christophe and {Wu}, Chunyang and {Wong}, Corinne and {Canton Ferrer}, Cristian and {Nikolaidis}, Cyrus and {Allonsius}, Damien and {Song}, Daniel and {Pintz}, Danielle and {Livshits}, Danny and {Wyatt}, Danny and {Esiobu}, David and {Choudhary}, Dhruv and {Mahajan}, Dhruv and {Garcia-Olano}, Diego and {Perino}, Diego and {Hupkes}, Dieuwke and {Lakomkin}, Egor and {AlBadawy}, Ehab and {Lobanova}, Elina and {Dinan}, Emily and {Smith}, Eric Michael and {Radenovic}, Filip and {Guzm{\'a}n}, Francisco and {Zhang}, Frank and {Synnaeve}, Gabriel and {Lee}, Gabrielle and {Anderson}, Georgia Lewis and {Thattai}, Govind and {Nail}, Graeme and {Mialon}, Gregoire and {Pang}, Guan and {Cucurell}, Guillem and {Nguyen}, Hailey and {Korevaar}, Hannah and {Xu}, Hu and {Touvron}, Hugo and {Zarov}, Iliyan and {Arrieta Ibarra}, Imanol and {Kloumann}, Isabel and {Misra}, Ishan and {Evtimov}, Ivan and {Zhang}, Jack and {Copet}, Jade and {Lee}, Jaewon and {Geffert}, Jan and {Vranes}, Jana and {Park}, Jason and {Mahadeokar}, Jay and {Shah}, Jeet and {van der Linde}, Jelmer and {Billock}, Jennifer and {Hong}, Jenny and {Lee}, Jenya and {Fu}, Jeremy and {Chi}, Jianfeng and {Huang}, Jianyu and {Liu}, Jiawen and {Wang}, Jie and {Yu}, Jiecao and {Bitton}, Joanna and {Spisak}, Joe and {Park}, Jongsoo and {Rocca}, Joseph and {Johnstun}, Joshua and {Saxe}, Joshua and {Jia}, Junteng and {Vasuden Alwala}, Kalyan and {Prasad}, Karthik and {Upasani}, Kartikeya and {Plawiak}, Kate and {Li}, Ke and {Heafield}, Kenneth and {Stone}, Kevin and {El-Arini}, Khalid and {Iyer}, Krithika and {Malik}, Kshitiz and {Chiu}, Kuenley and {Bhalla}, Kunal and {Lakhotia}, Kushal and {Rantala-Yeary}, Lauren and {van der Maaten}, Laurens and {Chen}, Lawrence and {Tan}, Liang and {Jenkins}, Liz and {Martin}, Louis and {Madaan}, Lovish and {Malo}, Lubo and {Blecher}, Lukas and {Landzaat}, Lukas and {de Oliveira}, Luke and {Muzzi}, Madeline and {Pasupuleti}, Mahesh and {Singh}, Mannat and {Paluri}, Manohar and {Kardas}, Marcin and {Tsimpoukelli}, Maria and {Oldham}, Mathew and {Rita}, Mathieu and {Pavlova}, Maya and {Kambadur}, Melanie and {Lewis}, Mike and {Si}, Min and {Singh}, Mitesh Kumar and {Hassan}, Mona and {Goyal}, Naman and {Torabi}, Narjes and {Bashlykov}, Nikolay and {Bogoychev}, Nikolay and {Chatterji}, Niladri and {Zhang}, Ning and {Duchenne}, Olivier and {{\c{C}}elebi}, Onur and {Alrassy}, Patrick and {Zhang}, Pengchuan and {Li}, Pengwei and {Vasic}, Petar and {Weng}, Peter and {Bhargava}, Prajjwal and {Dubal}, Pratik and {Krishnan}, Praveen and {Singh Koura}, Punit and {Xu}, Puxin and {He}, Qing and {Dong}, Qingxiao and {Srinivasan}, Ragavan and {Ganapathy}, Raj and {Calderer}, Ramon and {Silveira Cabral}, Ricardo and {Stojnic}, Robert and {Raileanu}, Roberta and {Maheswari}, Rohan and {Girdhar}, Rohit and {Patel}, Rohit and {Sauvestre}, Romain and {Polidoro}, Ronnie and {Sumbaly}, Roshan and {Taylor}, Ross and {Silva}, Ruan and {Hou}, Rui and {Wang}, Rui and {Hosseini}, Saghar and {Chennabasappa}, Sahana and {Singh}, Sanjay and {Bell}, Sean and {Kim}, Seohyun Sonia and {Edunov}, Sergey and {Nie}, Shaoliang and {Narang}, Sharan and {Raparthy}, Sharath and {Shen}, Sheng and {Wan}, Shengye and {Bhosale}, Shruti and {Zhang}, Shun and {Vandenhende}, Simon and {Batra}, Soumya and {Whitman}, Spencer and {Sootla}, Sten and {Collot}, Stephane and {Gururangan}, Suchin and {Borodinsky}, Sydney and {Herman}, Tamar and {Fowler}, Tara and {Sheasha}, Tarek and {Georgiou}, Thomas and {Scialom}, Thomas and {Speckbacher}, Tobias},
        title = "{The Llama 3 Herd of Models}",
      journal = {arXiv e-prints},
         year = 2024,
        month = jul,
          eid = {arXiv:2407.21783},
        pages = {arXiv:2407.21783},
          doi = {10.48550/arXiv.2407.21783},
archivePrefix = {arXiv},
       eprint = {2407.21783},
 primaryClass = {cs.AI},
       adsurl = {https://ui.adsabs.harvard.edu/abs/2024arXiv240721783G}
}

@article{abdin2024phi,
  title={Phi-4 technical report},
  author={Abdin, Marah and Aneja, Jyoti and Behl, Harkirat and Bubeck, S{\'e}bastien and Eldan, Ronen and Gunasekar, Suriya and Harrison, Michael and Hewett, Russell J and Javaheripi, Mojan and Kauffmann, Piero and others},
  journal={arXiv preprint arXiv:2412.08905},
  year={2024}
}

@article{comanici2025gemini,
  title={Gemini 2.5: Pushing the frontier with advanced reasoning, multimodality, long context, and next generation agentic capabilities},
  author={Comanici, Gheorghe and Bieber, Eric and Schaekermann, Mike and Pasupat, Ice and Sachdeva, Noveen and Dhillon, Inderjit and Blistein, Marcel and Ram, Ori and Zhang, Dan and Rosen, Evan and others},
  journal={arXiv preprint arXiv:2507.06261},
  year={2025}
}

@article{li2024automated,
  title={Automated Evaluation of Large Vision-Language Models on Self-driving Corner Cases},
  author={Li, Yanze and Zhang, Wenhua and Chen, Kai and Liu, Yanxin and Li, Pengxiang and Gao, Ruiyuan and Hong, Lanqing and Tian, Meng and Zhao, Xinhai and Li, Zhenguo and others},
  journal={arXiv preprint arXiv:2404.10595},
  year={2024}
  }

@inproceedings{qian2024nuscenes,
  title={Nuscenes-qa: A multi-modal visual question answering benchmark for autonomous driving scenario},
  author={Qian, Tianwen and Chen, Jingjing and Zhuo, Linhai and Jiao, Yang and Jiang, Yu-Gang},
  booktitle={Proceedings of the AAAI Conference on Artificial Intelligence},
  volume={38},
  number={5},
  pages={4542--4550},
  year={2024}
}

@article{guo2026surds,
  title={Surds: Benchmarking spatial understanding and reasoning in driving scenarios with vision language models},
  author={Guo, Xianda and Zhang, Ruijun and Duan, Yiqun and He, Yuhang and Nie, Dujun and Huang, Wenke and Zhang, Chenming and Liu, Shuai and Zhao, Hao and Chen, Long},
  journal={Advances in Neural Information Processing Systems},
  volume={38},
  year={2026}
}

@inproceedings{xie2025vlms,
  title={Are VLMs Ready for Autonomous Driving? An Empirical Study from the Reliability, Data and Metric Perspectives},
  author={Xie, Shaoyuan and Kong, Lingdong and Dong, Yuhao and Sima, Chonghao and Zhang, Wenwei and Chen, Qi Alfred and Liu, Ziwei and Pan, Liang},
  booktitle={Proceedings of the IEEE/CVF International Conference on Computer Vision},
  pages={6585--6597},
  year={2025}
}

@inproceedings{wang2025omnidrive,
  title={Omnidrive: A holistic vision-language dataset for autonomous driving with counterfactual reasoning},
  author={Wang, Shihao and Yu, Zhiding and Jiang, Xiaohui and Lan, Shiyi and Shi, Min and Chang, Nadine and Kautz, Jan and Li, Ying and Alvarez, Jose M},
  booktitle={Proceedings of the computer vision and pattern recognition conference},
  pages={22442--22452},
  year={2025}
}

@inproceedings{hu2023planning,
  title={Planning-oriented autonomous driving},
  author={Hu, Yihan and Yang, Jiazhi and Chen, Li and Li, Keyu and Sima, Chonghao and Zhu, Xizhou and Chai, Siqi and Du, Senyao and Lin, Tianwei and Wang, Wenhai and others},
  booktitle={Proceedings of the IEEE/CVF conference on computer vision and pattern recognition},
  pages={17853--17862},
  year={2023}
}

@article{li2022bevformer,
  title={Bevformer: Learning bird’s-eye-view representation from multi-camera images via spatiotemporal transformers.(2022)},
  author={Li, Zhiqi and Wang, Wenhai and Li, Hongyang and Xie, Enze and Sima, Chonghao and Lu, Tong and Yu, Qiao and Dai, Jifeng},
  journal={URL https://arxiv. org/abs/2203.17270},
  volume={10},
  year={2022}
}

@inproceedings{jiang2023vad,
  title={Vad: Vectorized scene representation for efficient autonomous driving},
  author={Jiang, Bo and Chen, Shaoyu and Xu, Qing and Liao, Bencheng and Chen, Jiajie and Zhou, Helong and Zhang, Qian and Liu, Wenyu and Huang, Chang and Wang, Xinggang},
  booktitle={Proceedings of the IEEE/CVF International Conference on Computer Vision},
  pages={8340--8350},
  year={2023}
}

@inproceedings{liao2025diffusiondrive,
  title={Diffusiondrive: Truncated diffusion model for end-to-end autonomous driving},
  author={Liao, Bencheng and Chen, Shaoyu and Yin, Haoran and Jiang, Bo and Wang, Cheng and Yan, Sixu and Zhang, Xinbang and Li, Xiangyu and Zhang, Ying and Zhang, Qian and others},
  booktitle={Proceedings of the Computer Vision and Pattern Recognition Conference},
  pages={12037--12047},
  year={2025}
}
\clearpage
\appendix
\section*{Appendix}
\section{Dataset Summary}\label{app:dataset-risk}

\subsection{Source Datasets and Usage Statistics}\label{app:datasets-usage}
Table~\ref{tab:app-dataset-usage} summarizes the three source datasets used in Drive-P2D, including their regions, scene types, task coverage, and the corresponding image/QA usage statistics.
\begin{table*}[t]
\centering
\caption{Source datasets and our usage in Drive-P2D.}
\label{tab:app-dataset-usage}
\setlength{\tabcolsep}{5pt}
\renewcommand{\arraystretch}{1.05}
\small
\begin{tabular}{l p{2.6cm} p{5.0cm} p{2.0cm} c c}
\toprule
Dataset & Country / Region & Scene Types & Covered Tasks & Used Images & QA Items \\
\midrule
nuScenes & Boston; Singapore &
Urban streets, dense traffic, diverse maneuvers, diverse weather &
\begin{tabular}[t]{@{}l@{}}
Object-1\\
Object-2\\
Decision-1\\
Decision-2
\end{tabular} &
465 & 1910 \\
BDD100K & United States &
Residential areas, highways, city streets, parking lots, gas stations, tunnels &
\begin{tabular}[t]{@{}l@{}}
Object-1\\
Object-2\\
Scene-1\\
Scene-2\\
Decision-1\\
Decision-2
\end{tabular} &
450 & 3220 \\
KITTI & Karlsruhe, Germany and surrounding areas &
Urban streets, residential areas, campus, rural roads, highways &
\begin{tabular}[t]{@{}l@{}}
Object-1\\
Object-2\\
Decision-1\\
Decision-2
\end{tabular} &
380 & 1520 \\
\bottomrule
\end{tabular}
\end{table*}

\noindent\textbf{Total images used:} 1295.\quad
\textbf{Total QA items:} 6650.

\subsection{Design Intentions of the Six Tasks}\label{app:task-intentions}
Table~\ref{tab:app-task-intentions} summarizes the design intention of each task in Drive-P2D.

\begin{table*}[t]
\centering
\caption{Design intentions of the six tasks in Drive-P2D.}
\label{tab:app-task-intentions}
\setlength{\tabcolsep}{10pt}
\renewcommand{\arraystretch}{1.12}
\small
\begin{tabular}{p{0.22\textwidth} p{0.73\textwidth}}
\toprule
Task & Design intention \\
\midrule
Object-1 & Identifies the key object, serving as the starting point of the model’s reasoning process, thereby testing its ability to capture task-relevant information and filter out irrelevant information. \\
Object-2 & Evaluates the model’s inferential capacity, requiring it to infer the object’s state from visual observations, thereby constraining the feasible subsequent actions. \\
Scene-1 & Captures the overall environmental context, providing an overview of general driving conditions rather than frame-specific operational requirements (e.g., rainy or low-light conditions that affect safe driving speed). \\
Scene-2 & Extends critical information extraction from object-level perception to the broader scene context, imposing additional constraints to support downstream decision-making. \\
Decision-1 & Represents the terminal point of the reasoning process, measuring the model’s core ability to make correct decisions. \\
Decision-2 & Complements Decision-1 by evaluating responses to suboptimal options, providing a more comprehensive evaluation of VLMs’ capability boundaries. \\
\bottomrule
\end{tabular}
\end{table*}

\subsection{Risk Rating Rules}\label{app:risk-rubric}
We define scenario risk as an ordinal scale that measures the short-term likelihood of a collision and the potential severity of its consequences under a reasonable human-driver policy. The rating is based on safety-relevant cues visible in a single image (e.g., relative positions, right-of-way, vulnerable road users, and abnormal road conditions).

Two annotators independently assign an integer level $r \in \{1,2,3,4,5\}$ according to the rules in Table~\ref{tab:app-risk-rules}. We take the mean score $\bar{r}$ as the final risk score and define the high-risk split as $\bar{r} \geq 4.0$.

\begin{table*}[t]
\centering
\caption{Five-level risk rules for scenario annotation.}
\label{tab:app-risk-rules}
\setlength{\tabcolsep}{6pt}
\renewcommand{\arraystretch}{1.08}
\small
\begin{tabular}{c p{0.88\textwidth}}
\toprule
Level & Operational definition (single-image, near-term driving risk) \\
\midrule
1 & \textbf{Minimal:} No safety-critical risk is observed; no immediate action is required beyond lane keeping and maintaining speed. \\
2 & \textbf{Low:} The scene shows mild complexity or weak risk cues; a cautious driver may reduce speed or increase headway, but immediate intervention is typically unnecessary. \\
3 & \textbf{Moderate:} A conflict may arise in the near term; the driver should increase attention and be ready to brake or yield when needed, and the situation is likely to be resolved safely with timely response. \\
4 & \textbf{High:} Strong risk cues are present and an unsafe outcome is likely without timely action; the driver should take an immediate safety action (e.g., braking or yielding). \\
5 & \textbf{Severe:} A safety-critical situation with near-collision cues; the likelihood of a collision is very high (e.g., a pedestrian suddenly appears within the field of view at close range). \\
\bottomrule
\end{tabular}
\end{table*}


\subsection{Similarity Illustration}\label{app:similar}

\begin{figure*}[t]
  \centering
  \includegraphics[width=\textwidth]{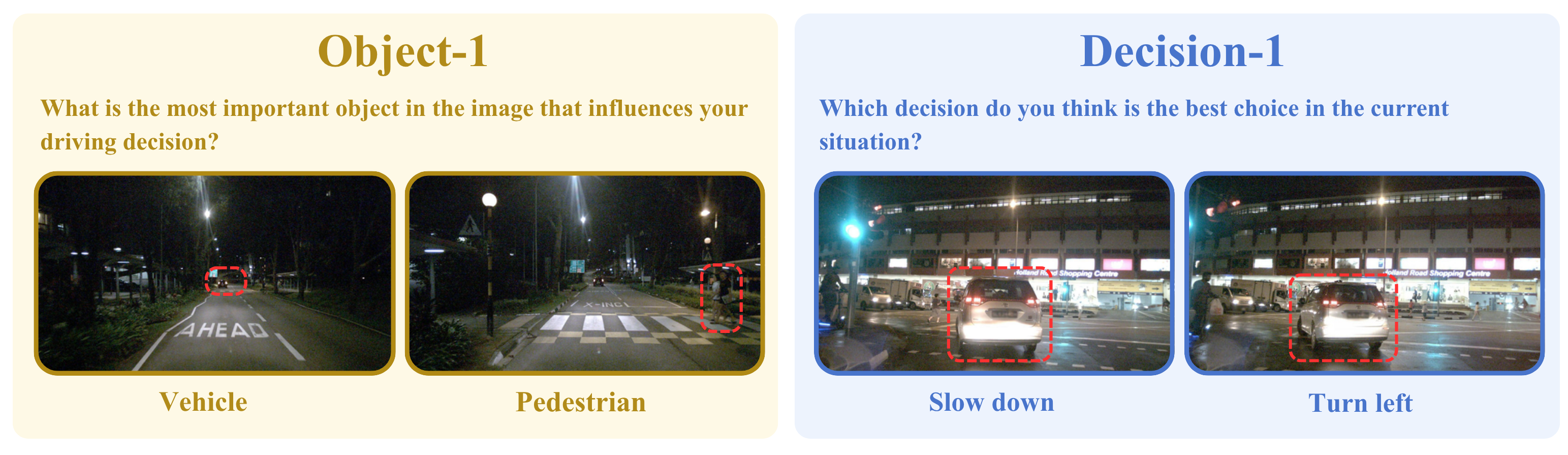}
  \caption{Examples of similar-scene robustness test.}
  \label{fig:app-similarity}
\end{figure*}

This section illustrates the curated near-duplicate scene pairs used to evaluate robustness. The pairs are selected by high embedding similarity and are intended to test whether models maintain causally consistent decisions under small visual changes. Figure~\ref{fig:app-similarity} presents representative pairs.

\section{Error-Mode Taxonomy \& Examples \& Analyzer Model}
\subsection{Error-Mode Taxonomy and Examples}\label{app:exp-taxonomy}
We use the following taxonomy to analyze reasoning and characterize failure modes.

\paragraph{Error-Mode Taxonomy.}
\begin{enumerate}[leftmargin=1.5em, itemsep=2pt, topsep=2pt]
\item \textbf{Logical Reasoning Error}: Perception may be correct, but the subsequent reasoning is logically invalid or contradicts traffic rules or causality.
\item \textbf{Semantic Feature Omission}: The model overlooks or misjudges safety-critical semantic/visual cues (e.g., brake lights, turn signals, pedestrian gestures) or other discriminative attributes necessary for correct recognition and categorization.
\item \textbf{Model Hallucination}: The model produces information inconsistent with the actual input, e.g., inventing non-existent objects, attributes, or relations.
\item \textbf{Modality Imbalance}: The model over-relies on one modality while neglecting information from another (e.g., text vs. image).
\item \textbf{Spatial Relation Misjudgment}: The model makes incorrect judgments about spatial relations (e.g., distance, relative position, depth).
\item \textbf{Limited Logical Inference}: The model draws decisions from insufficient local evidence without integrating broader scene context.
\item \textbf{Generalization Deficit}: The model generalizes poorly to rare or out-of-distribution scenarios (e.g., construction zones, traffic incidents, temporary lane closures).
\item \textbf{Decision Boundary Instability}: Near critical operating points, small input perturbations cause large changes in the chosen action.
\item \textbf{Target Priority Misjudgment}: The model fails to correctly prioritize among competing objectives (e.g., yielding to pedestrians vs. maintaining speed).
\end{enumerate}

\begin{figure*}[t]
  \centering
  \includegraphics[width=\textwidth]{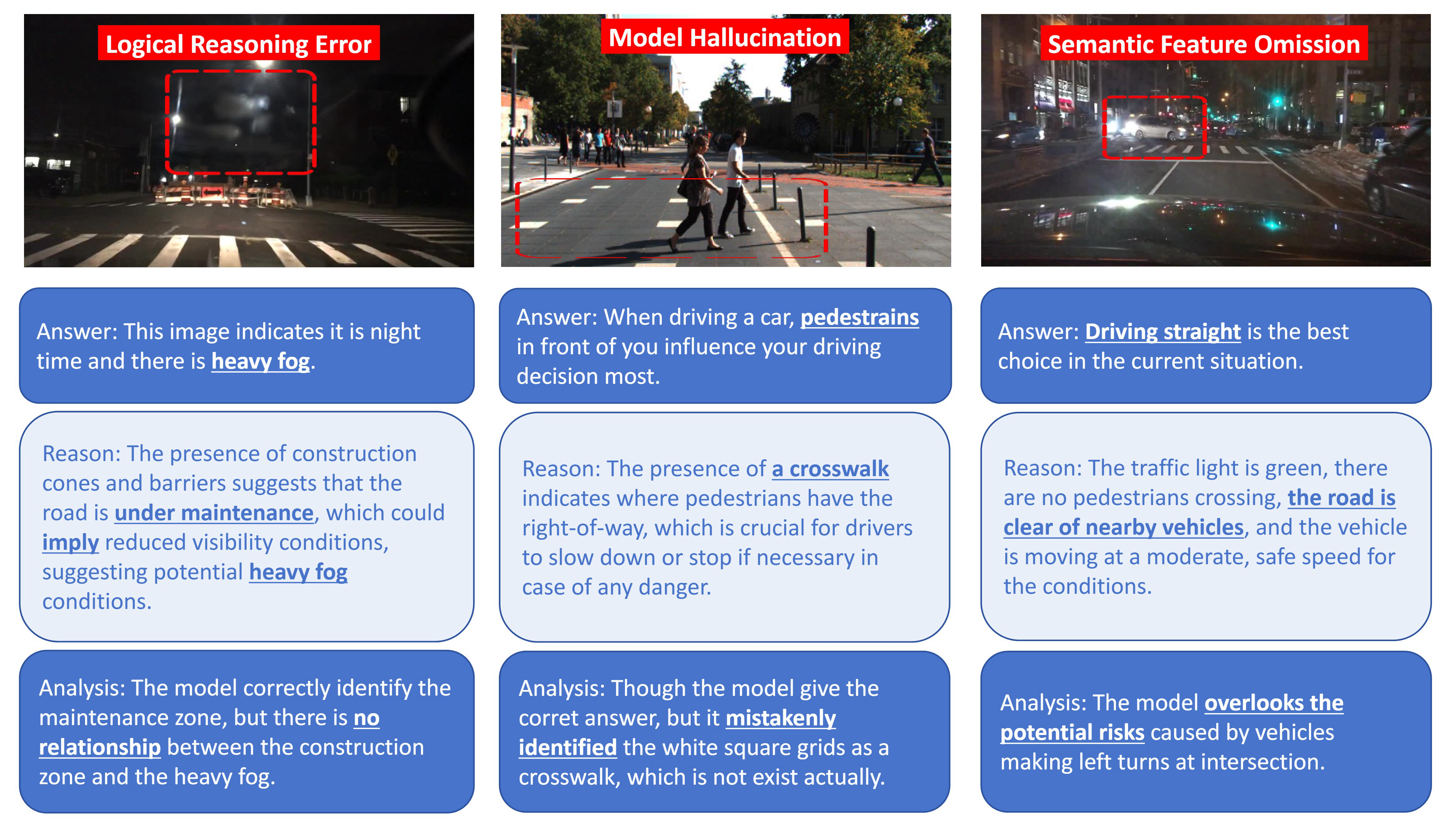}
  \caption{Representative Error Examples in Error-Mode Analysis}
  \label{expla}
\end{figure*}

Figure~\ref{expla} provides representative, annotated examples for some categories.

\subsection{Analyzer Model I/O Schematic \& Training}\label{app:exp-io}
The analyzer consumes an image, question, options, the ground-truth answer, and a model’s reasoning and final answer, and outputs one or more predicted error labels from the taxonomy. See Figure~\ref{fig:exp-system-user-prompt} for the system instruction and user input template used in our analysis.

For analyzer training, we use Qwen2.5-VL-7B-Instruct as the base model and perform supervised fine-tuning for this classification task.
Training labels are obtained through manual annotation.
During fine-tuning, we freeze the image encoder and multimodal projector, train only the LLM, and use a learning rate of 1e-4 for 5 epochs.

\section{Experimental Details}
This section lists full model names, the repetition protocol, and the prompts we used. The models evaluated in this work include the InternVL2.5 family \citep{chen2024internvl}, GPT-4.1 \citep{achiam2023gpt}, Gemini-2.5-Pro \citep{comanici2025gemini}, Gemma-3 \citep{gemmateam2025gemma3technicalreport}, Llama-3.2 \citep{2024arXiv240721783G}, Llava \citep{liu2023improved}, Phi-4 \citep{abdin2024phi}, and Qwen2.5-VL \citep{bai2025qwen25vltechnicalreport}.

\subsection{Experimental Setup}\label{app:exper_setup}

The benchmark assesses VLMs with an image, a question, and answer options; Object and Decision tasks are single-choice, whereas Scene tasks are multiple-choice. Models are required to respond with the specified format. For multiple-choice questions, full credit is awarded for exact matches, partial credit for subsets. Selection of any incorrect options results in a zero score. We abbreviate model names by capitalizing only the first letter, and the full-name mapping is provided in Appendix~\ref{app:model-full-names}.
For all evaluated VLMs, we used the official model checkpoints or APIs. Unless otherwise specified, we used each model's default inference configuration. 

\subsection{Model List and Full Names}\label{app:model-full-names}
\begin{itemize}[leftmargin=1.2em, itemsep=0.4em]
  \item \textbf{Gemma (4B)} $\rightarrow$ gemma-3-4b-it
  \item \textbf{Gemma (27B)} $\rightarrow$ gemma-3-27b-it
  \item \textbf{Llama (11B)} $\rightarrow$ Llama-3.2-11B-Vision-Instruct
  \item \textbf{Llama (90B)} $\rightarrow$ Llama-3.2-90B-Vision-Instruct
  \item \textbf{Llava (7B)} $\rightarrow$ LLaVA-v1.6-Mistral-7B-hf
  \item \textbf{Llava (72B)} $\rightarrow$ LLaVA-NeXT-72B-hf
  \item \textbf{Phi (6B)} $\rightarrow$ Phi-4-multimodal-instruct
  \item \textbf{Qwen (7B)} $\rightarrow$ Qwen2.5-VL-7B-Instruct
  \item \textbf{Qwen (72B)} $\rightarrow$ Qwen2.5-VL-72B-Instruct
  \item \textbf{InternVL (1B)} $\rightarrow$ InternVL2.5-1B-MPO
  \item \textbf{InternVL (2B)} $\rightarrow$ InternVL2.5-2B-MPO
  \item \textbf{InternVL (4B)} $\rightarrow$ InternVL2.5-4B-MPO
  \item \textbf{InternVL (8B)} $\rightarrow$ InternVL2.5-8B-MPO
  \item \textbf{InternVL (26B)} $\rightarrow$ InternVL2.5-26B-MPO
  \item \textbf{InternVL (38B)} $\rightarrow$ InternVL2.5-38B-MPO
  \item \textbf{InternVL (78B)} $\rightarrow$ InternVL2.5-78B-MPO
\end{itemize}

\subsection{Repetition Protocol}
We run each configuration twice and report the average across runs. For open-source models, inference was performed on NVIDIA A100 GPUs; for closed-source models, inference was conducted via the vendors' official API endpoints.

\subsection{Benchmark Prompts (System/User)}
Figure~\ref{fig:cot-system-user-prompt} shows the prompt templates for the single-choice and multiple-choice reasoning-and-answer prompts used in our experiments.

\begin{figure*}[!t]
  \centering
  \includegraphics[width=\linewidth]{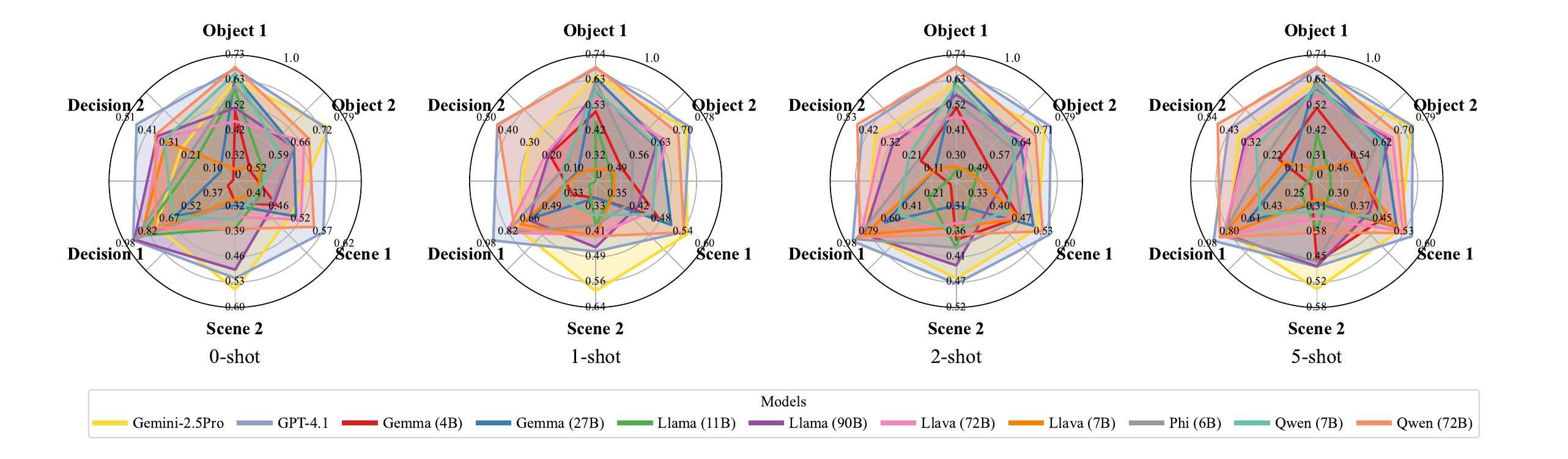}
  \caption{Model performance on all-scenarios under 0/1/2/5-shot prompting}
  \label{fig:app-all-radar}
\end{figure*}

\begin{figure*}[!t]
  \centering
  \includegraphics[width=\linewidth]{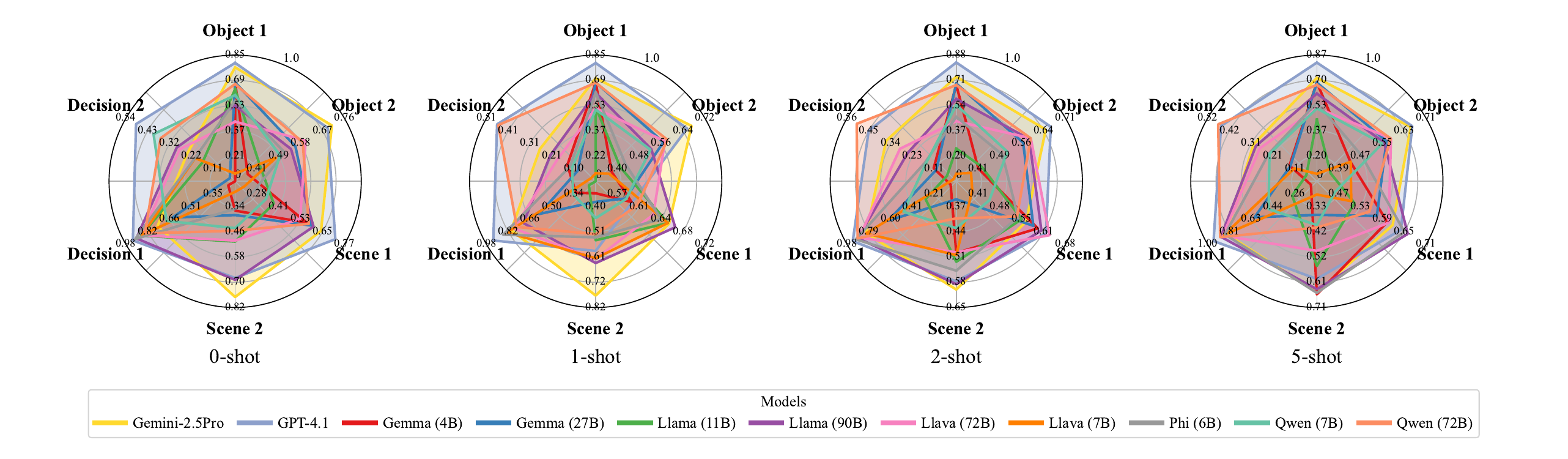}
  \caption{Model performance on high-risk-scenarios under 0/1/2/5-shot prompting}
  \label{fig:high-risk-all-radar}
\end{figure*}

\subsection{Annotation Templates for the Six Tasks}\label{app:annotation-templates}
\begin{figure}[t]
  \centering
  \fbox{%
  \begin{minipage}{\dimexpr\columnwidth-2\fboxsep-2\fboxrule\relax}
    \centering
    \small
    \textbf{Example Image}\par
    \medskip\hrule\medskip
    \includegraphics[width=0.92\linewidth]{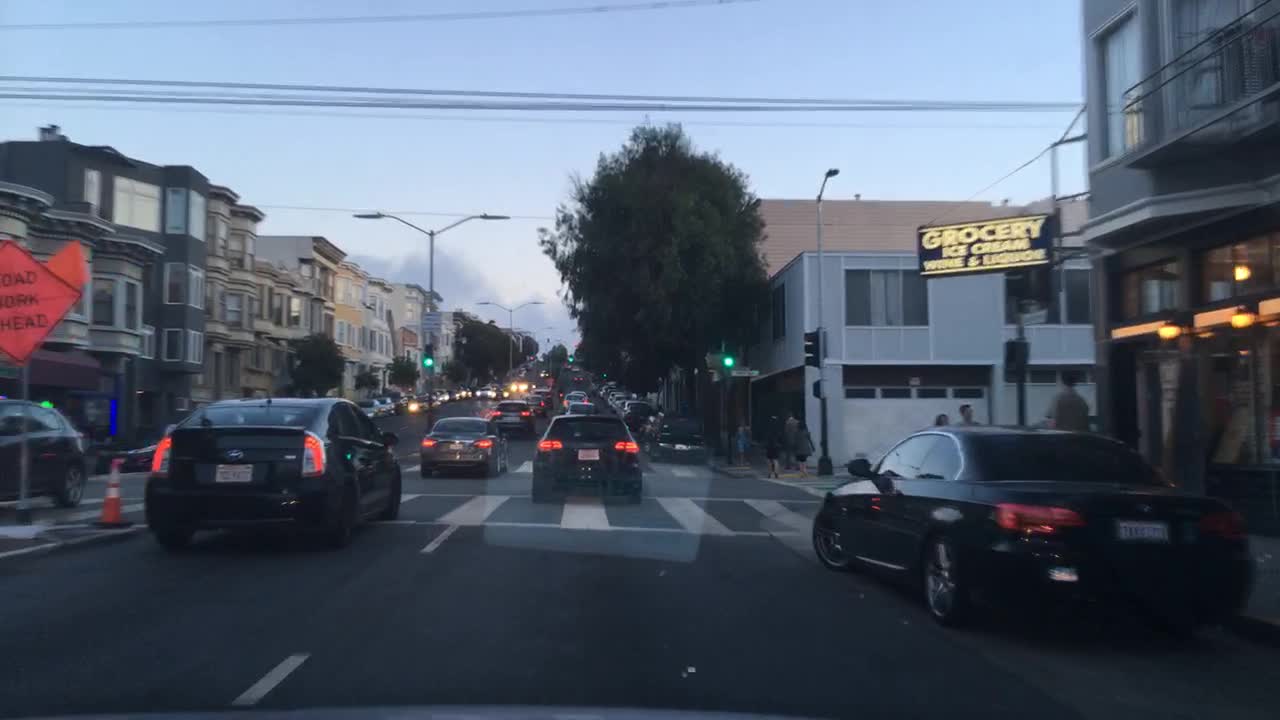}
  \end{minipage}
  }
  \caption{Example image used for the six-task annotation templates.}
  \label{fig:app-annotation-example-image}
\end{figure}

We provide a unified labeling template to ensure consistent annotation across the six Drive-P2D tasks.
Figure~\ref{fig:app-annotation-example-image} shows the example image, and Figure~\ref{fig:app-annotation-template} presents its scenario-level risk label (\texttt{danger\_score}) and question--option--answer templates.
All \uline{underlined} fields are placeholders to be replaced according to the objects, states, and scene conditions in each image.

\begin{figure*}[t]
  \centering
  \setlength{\fboxsep}{8pt}

  \fbox{%
  \begin{minipage}{0.965\textwidth}
    \small
    \textbf{Danger Score Label}\par
    \medskip\hrule\medskip
    {\ttfamily\raggedright\setlength{\parindent}{0pt}\setlength{\parskip}{3pt}%
    \begin{tabular}{@{}p{\linewidth}@{}}
\textbf{danger\_score}: \uline{2} \\
    \end{tabular}%
    }
  \end{minipage}
  }

  \vspace{1.0ex}

  \fbox{%
  \begin{minipage}{0.965\textwidth}
    \small
    \textbf{Object-1: Key object influencing the driving decision}\par
    \medskip\hrule\medskip
    {\ttfamily\raggedright\setlength{\parindent}{0pt}\setlength{\parskip}{3pt}%
    \begin{tabular}{@{}p{\linewidth}@{}}
QUESTION: When driving a car, what is the most important object in the image that influences your driving decision? Please select only one option.\\
OPTIONS: \uline{A.Building; B.Vehicle; C.Traffic light; D.Pedestrian; E.Street lamp}\\
ANSWER: \uline{C}
    \end{tabular}%
    }
  \end{minipage}
  }

  \vspace{0.9ex}

  \fbox{%
  \begin{minipage}{0.965\textwidth}
    \small
    \textbf{Object-2: State of the designated key object}\par
    \medskip\hrule\medskip
    {\ttfamily\raggedright\setlength{\parindent}{0pt}\setlength{\parskip}{3pt}%
    \begin{tabular}{@{}p{\linewidth}@{}}
QUESTION: What is the state of the \uline{traffic light}? Please select only one option.\\
OPTIONS: \uline{A.Green; B.Yellow; C.Red; D.There is no traffic light}\\
ANSWER: \uline{A}
    \end{tabular}%
    }
  \end{minipage}
  }

  \vspace{0.9ex}

  \fbox{%
  \begin{minipage}{0.965\textwidth}
    \small
    \textbf{Scene-1: Weather / illumination conditions}\par
    \medskip\hrule\medskip
    {\ttfamily\raggedright\setlength{\parindent}{0pt}\setlength{\parskip}{3pt}%
    \begin{tabular}{@{}p{\linewidth}@{}}
QUESTION: What is the weather condition in the image? Please select one or more options.\\
OPTIONS: A.Daytime; B.Nighttime; C.Rain; D.Snow; E.Heavy fog\\
ANSWER: \uline{A}
    \end{tabular}%
    }
  \end{minipage}
  }

  \vspace{0.9ex}

  \fbox{%
  \begin{minipage}{0.965\textwidth}
    \small
    \textbf{Scene-2: Special scene factors affecting decisions}\par
    \medskip\hrule\medskip
    {\ttfamily\raggedright\setlength{\parindent}{0pt}\setlength{\parskip}{3pt}%
    \begin{tabular}{@{}p{\linewidth}@{}}
QUESTION: What special scenes in the image could potentially affect the decision-making of the host vehicle? Please select one or more options.\\
OPTIONS: \uline{A.Heavy Traffic; B.Accident Scene; C.Construction Zone; D.Vehicle breakdown; E.School road section; F.Rain or Snow Weather; G.None of the above}\\
ANSWER: \uline{A}
    \end{tabular}%
    }
  \end{minipage}
  }
  
  \fbox{%
  \begin{minipage}{0.965\textwidth}
    \small
    \textbf{Decision-1: Optimal driving action}\par
    \medskip\hrule\medskip
    {\ttfamily\raggedright\setlength{\parindent}{0pt}\setlength{\parskip}{3pt}%
    \begin{tabular}{@{}p{\linewidth}@{}}
QUESTION: When driving a car, which decision do you think is the best choice in the current situation? Please select only one option.\\
OPTIONS: \uline{A.Drive straight; B.Slow down/make an emergency stop; C.Turn left; D.Turn right; E.Change lanes; F.Stay still}\\
ANSWER: \uline{A}
    \end{tabular}%
    }
  \end{minipage}
  }

  \vspace{0.9ex}

  \fbox{%
  \begin{minipage}{0.965\textwidth}
    \small
    \textbf{Decision-2: Risk of a specified action}\par
    \medskip\hrule\medskip
    {\ttfamily\raggedright\setlength{\parindent}{0pt}\setlength{\parskip}{3pt}%
    \begin{tabular}{@{}p{\linewidth}@{}}
QUESTION: If you \uline{speed up}, what level of risk will be faced? Please select only one option.\\
OPTIONS: A.Absolutely safe; B.Moderate risk; C.Extreme risk\\
ANSWER: \uline{B}
    \end{tabular}%
    }
  \end{minipage}
  }

  \vspace{0.9ex}

  \caption{Annotation templates instantiated on an example image. Underlined fields (\uline{...}) are placeholders that annotators replace according to the image.}
  \label{fig:app-annotation-template}
\end{figure*}

\begin{figure*}[t]
  \centering
  \setlength{\fboxsep}{8pt}

  \fbox{%
  \begin{minipage}{0.965\textwidth}
    \small
    \textbf{Prompt 1: Single-choice question --- System Instruction}\par
    \medskip\hrule\medskip
    {\ttfamily\raggedright\setlength{\parindent}{0pt}\setlength{\parskip}{4pt}%
    \begin{tabular}{@{}p{\linewidth}@{}}
You are an AI assistant specializing in automatic driving scene judgment. Your primary task is to analyze a given traffic scene and make a decision or evaluation based on it.\\[4pt]
\textbf{Input:}\\
You will be provided with an image, a corresponding question, and a set of possible options.\\[4pt]
\textbf{Answering Guidelines:}\\
\textbullet\ Your reasoning must be based on the visual evidence in the image and fundamental principles of safe driving.\\
\textbullet\ Analyze the situation step-by-step before making a final decision.\\[4pt]
\textbf{Instruction:}\\
You should choose one option from the given choices. First provide your brief reasoning as the model-generated textual justification, then put only the selected option letter in the \texttt{\textless answer\textgreater...\textless/answer\textgreater} field. Do not include any text outside the required tags.\\[4pt]
\textbf{Output Format:}\\
Your response must follow a two-part reasoning-and-answer format within a single turn. First, generate reasoning enclosed within \texttt{\textless think\textgreater...\textless/think\textgreater} tags; second, provide the final choice answer enclosed within \texttt{\textless answer\textgreater...\textless/answer\textgreater} tags. The structure must be exactly:\\[4pt]
\texttt{\textless think\textgreater}Your detailed reasoning process here.\texttt{\textless/think\textgreater}\texttt{\textless answer\textgreater}Your chosen option letter(s) here.\texttt{\textless/answer\textgreater}
    \end{tabular}%
    }
    \medskip\hrule
  \end{minipage}
  }

  \vspace{0.9ex}

  \fbox{%
  \begin{minipage}{0.965\textwidth}
    \small
    \textbf{Prompt 1: Single-choice question --- User Input Template}\par
    \medskip\hrule\medskip
    {\ttfamily\raggedright\setlength{\parindent}{0pt}\setlength{\parskip}{2pt}%
    \begin{tabular}{@{}p{\linewidth}@{}}
IMAGE: \{image\}\\
QUESTION: \{question\}\\
OPTIONS: \{options\}\\
    \end{tabular}%
    }
    \medskip\hrule
  \end{minipage}
  }

  \vspace{1.5em}

  \fbox{%
  \begin{minipage}{0.965\textwidth}
    \small
    \textbf{Prompt 2: Multiple-choice question --- System Instruction}\par
    \medskip\hrule\medskip
    {\ttfamily\raggedright\setlength{\parindent}{0pt}\setlength{\parskip}{4pt}%
    \begin{tabular}{@{}p{\linewidth}@{}}
You are an AI assistant specializing in automatic driving scene judgment. Your primary task is to analyze a given traffic scene and make a decision or evaluation based on it.\\[4pt]
\textbf{Input:}\\
You will be provided with an image, a corresponding question, and a set of possible options.\\[4pt]
\textbf{Answering Guidelines:}\\
\textbullet\ Your reasoning must be based on the visual evidence in the image and fundamental principles of safe driving.\\
\textbullet\ Analyze the situation step-by-step before making a final decision.\\[4pt]
\textbf{INSTRUCTION:}\\
You should select one or more options from the given choices. First provide your brief reasoning as the model-generated textual justification, then put only the selected option letter(s), separated by commas, in the \texttt{\textless answer\textgreater...\textless/answer\textgreater} field. Do not include any text outside the required tags.\\[4pt]
\textbf{Output Format:}\\
Your response must follow a two-part reasoning-and-answer format within a single turn. First, generate reasoning enclosed within \texttt{\textless think\textgreater...\textless/think\textgreater} tags; second, provide the final choice answer enclosed within \texttt{\textless answer\textgreater...\textless/answer\textgreater} tags. The structure must be exactly:\\[4pt]
\texttt{\textless think\textgreater}Your detailed reasoning process here.\texttt{\textless/think\textgreater}\texttt{\textless answer\textgreater}Your chosen option letter(s) here (comma-separated).\texttt{\textless/answer\textgreater}
    \end{tabular}%
    }
    \medskip\hrule
  \end{minipage}
  }

  \vspace{0.9ex}

  \fbox{%
  \begin{minipage}{0.965\textwidth}
    \small
    \textbf{Prompt 2: Multiple-choice question --- User Input Template}\par
    \medskip\hrule\medskip
    {\ttfamily\raggedright\setlength{\parindent}{0pt}\setlength{\parskip}{2pt}%
    \begin{tabular}{@{}p{\linewidth}@{}}
IMAGE: \{image\}\\
QUESTION: \{question\}\\
OPTIONS: \{options\}\\
    \end{tabular}%
    }
    \medskip\hrule
  \end{minipage}
  }

  \caption{System instruction and user input templates for the single-choice and multiple-choice reasoning-and-answer prompts.}
  \label{fig:cot-system-user-prompt}
\end{figure*}

\begin{figure*}[t]
  \centering
  \setlength{\fboxsep}{8pt}

  \fbox{%
  \begin{minipage}{0.965\textwidth}
    \small
    \textbf{Prompt 3: Error Tagging (Error-Mode Analysis) --- System Instruction}\par
    \medskip\hrule\medskip
    {\ttfamily\raggedright\setlength{\parindent}{0pt}\setlength{\parskip}{4pt}%
    \begin{tabular}{@{}p{\linewidth}@{}}
You are an autonomous-driving evaluation assistant for multimodal (image + text) questions. Your job is to analyze a model’s final answer and model-generated reasoning, detect which error patterns in the reasoning apply, and map them to the integer labels defined below. The user input will always contain: (1) the original question, (2) the question options, (3) the correct answer, (4) the model’s final answer, (5) the model-generated reasoning, and (6) an image associated with the question. Use both the image and the textual inputs (final answer + reasoning) to decide which label(s) apply.\\[4pt]
\textbf{Label definitions (0--9):}\\
0 = Correct — The model’s final answer is correct and its reasoning is sound and sufficient.\\
1 = Logical Reasoning Error — Perception may be correct, but the subsequent reasoning is logically invalid or contradicts traffic rules or causality.\\
2 = Semantic Feature Omission — The model overlooks or misjudges safety-critical semantic/visual cues (e.g., brake lights, turn signals, pedestrian gestures) or other discriminative attributes necessary for correct recognition and categorization.\\
3 = Model Hallucination — The model produces information inconsistent with the actual input (e.g., invents non-existent objects, attributes, or relations).\\
4 = Modality Imbalance — The model over-relies on one modality (text or image) while neglecting the other, causing an error.\\
5 = Spatial Relation Misjudgment — The model makes incorrect judgments about spatial relations (e.g., distance, relative position, depth).\\
6 = Limited Logical Inference — The model derives a decision from insufficient local evidence without integrating broader scene context.\\
7 = Generalization Deficit — The model generalizes poorly to rare or out-of-distribution scenarios (e.g., construction zones, traffic incidents, temporary lane closures).\\
8 = Decision Boundary Instability — Near critical operating points, small input perturbations cause large changes in the chosen action.\\
9 = Target Priority Misjudgment — The model fails to correctly prioritize among competing objectives (e.g., yielding to pedestrians vs. maintaining speed).\\[4pt]
\textbf{Classification rules:}\\
\textbullet\ Use both the image and the provided text (final answer + reasoning) to decide labels.\\
\textbullet\ If the final answer is correct but the reasoning is flawed in any way, do NOT output 0; instead output the label(s) for the flawed aspects (you may include multiple labels).\\[4pt]
\textbf{Output format:}\\
Your output must be ONLY a comma-separated list of integers (no words, no punctuation except commas and digits). Examples:\\
0\\
7\\
1,3\\
2,6,9\\
    \end{tabular}%
    }
    \medskip\hrule
  \end{minipage}
  }

  \vspace{0.9ex}

  \fbox{%
  \begin{minipage}{0.965\textwidth}
    \small
    \textbf{Prompt 3: Error Tagging (Error-Mode Analysis) --- User Input Template}\par
    \medskip\hrule\medskip
    {\ttfamily\raggedright\setlength{\parindent}{0pt}\setlength{\parskip}{2pt}%
    \begin{tabular}{@{}p{\linewidth}@{}}
IMAGE: \{image\}\\
QUESTION: \{question\}\\
OPTIONS: \{options\}\\
ANSWER: \{answer\}\\
MODEL FINAL ANSWER: \{model\_answer\}\\
MODEL REASONING: \{reasoning\}\\
    \end{tabular}%
    }
    \medskip\hrule
  \end{minipage}
  }

  \caption{System instruction and user input template for the error-mode tagging prompt.}
  \label{fig:exp-system-user-prompt}
\end{figure*}

\begin{table*}[t]
\centering
\caption{Evaluation on High-Risk Scenarios (0-shot)}
\label{tab:app-eval-highrisk-intern-excluded}
\setlength{\tabcolsep}{6pt}\renewcommand{\arraystretch}{1.05}\small
\begin{tabular}{lccccccccc}
\toprule
Model & Size & Open & Object-1 & Object-2 & Scene-1 & Scene-2 & Decision-1 & Decision-2 & Avg Score \\
\midrule
Gemini-2.5-Pro & - & \xmark & 77.40 & \textbf{\underline{70.62}} & 75.00 & 23.68 & 63.84 & \textbf{\underline{77.40}} & 64.66 \\
GPT-4.1 & - & \xmark & \textbf{\underline{80.23}} & 68.93 & \textbf{\underline{92.76}} & \textbf{\underline{48.90}} & \textbf{\underline{71.75}} & 68.36 & \textbf{\underline{71.82}} \\
\midrule
Gemma & 4B & \cmark & 67.62 & 37.05 & 24.14 & \phantom{0}0.00 & 55.88 & 36.34 & 36.84 \\
Gemma & 27B & \cmark & 67.74 & 56.11 & 64.37 & \phantom{0}2.64 & 58.78 & 38.43 & 48.01 \\
Llama & 11B & \cmark & 63.37 & 42.08 & 86.82 & 16.13 & 37.70 & 51.04 & 49.52 \\
Llama & 90B & \cmark & 53.04 & 54.33 & 90.25 & 28.83 & 59.71 & 69.26 & 59.24 \\
Llava & 7B & \cmark & \phantom{0}9.37 & 48.37 & 86.02 & 19.78 & 21.37 & 27.37 & 35.38 \\
Llava & 72B & \cmark & 42.57 & 60.19 & 86.42 & 27.37 & 52.28 & 50.56 & 53.23 \\
Phi & 6B & \cmark & 55.89 & 51.88 & 91.13 & 19.87 & 26.36 & 35.48 & 46.77 \\
Qwen & 7B & \cmark & 59.57 & 49.97 & 68.55 & 40.27 & 33.42 & 44.63 & 49.40 \\
Qwen & 72B & \cmark & 66.69 & 59.01 & 84.09 & 36.78 & 57.06 & 45.87 & 58.25 \\
\bottomrule
\end{tabular}
\end{table*}
\section{Supplemental Figures and Tables}
We provide correlation heatmaps, radar plots, and per-shot score tables to complement the main results and support fine-grained inspection.

\subsection{High-Risk Scenario Results (0-shot)}\label{app:highrisk-0shot}
Table~\ref{tab:app-eval-highrisk-intern-excluded} reports the 0-shot performance of all evaluated models on the high-risk split.

\subsection{Radar Charts}\label{app:radar-charts}
Figures~\ref{fig:app-all-radar} and~\ref{fig:high-risk-all-radar} summarize shot-wise performance trends and non-monotonic behaviors under few-shot prompting, for the all-scenario and high-risk splits, respectively.

\begin{table*}[t]
\centering
\caption{Analyzer robustness across source models and risk strata}
\label{tab:app-analyzer-robustness}
\setlength{\tabcolsep}{6pt}
\renewcommand{\arraystretch}{1.05}
\small
\begin{tabular}{lcccc}
\toprule
Group & Exact Match $\uparrow$ & Partial Match & Mismatch $\downarrow$ & Avg Score $\uparrow$ \\
\midrule
Average & \textbf{64.79} & \textbf{13.33} & \textbf{21.88} & \textbf{71.46} \\
InternVL (high-risk) & 70.00 & 13.75 & 16.25 & 76.88 \\
InternVL (non-high-risk) & 62.50 & 12.50 & 25.00 & 68.75 \\
Qwen (high-risk) & 66.25 & 13.75 & 20.00 & 73.12 \\
Qwen (non-high-risk) & 63.75 & 10.00 & 26.25 & 68.75 \\
GPT-4.1 (high-risk) & 65.00 & 16.25 & 18.75 & 73.12 \\
GPT-4.1 (non-high-risk) & 61.25 & 13.75 & 25.00 & 68.12 \\
\bottomrule
\end{tabular}
\end{table*}

\subsection{Correlation Matrices (Qwen Total)}\label{app:corr-qwen-total}
\begin{figure*}[t]
  \centering
  \includegraphics[width=\textwidth]{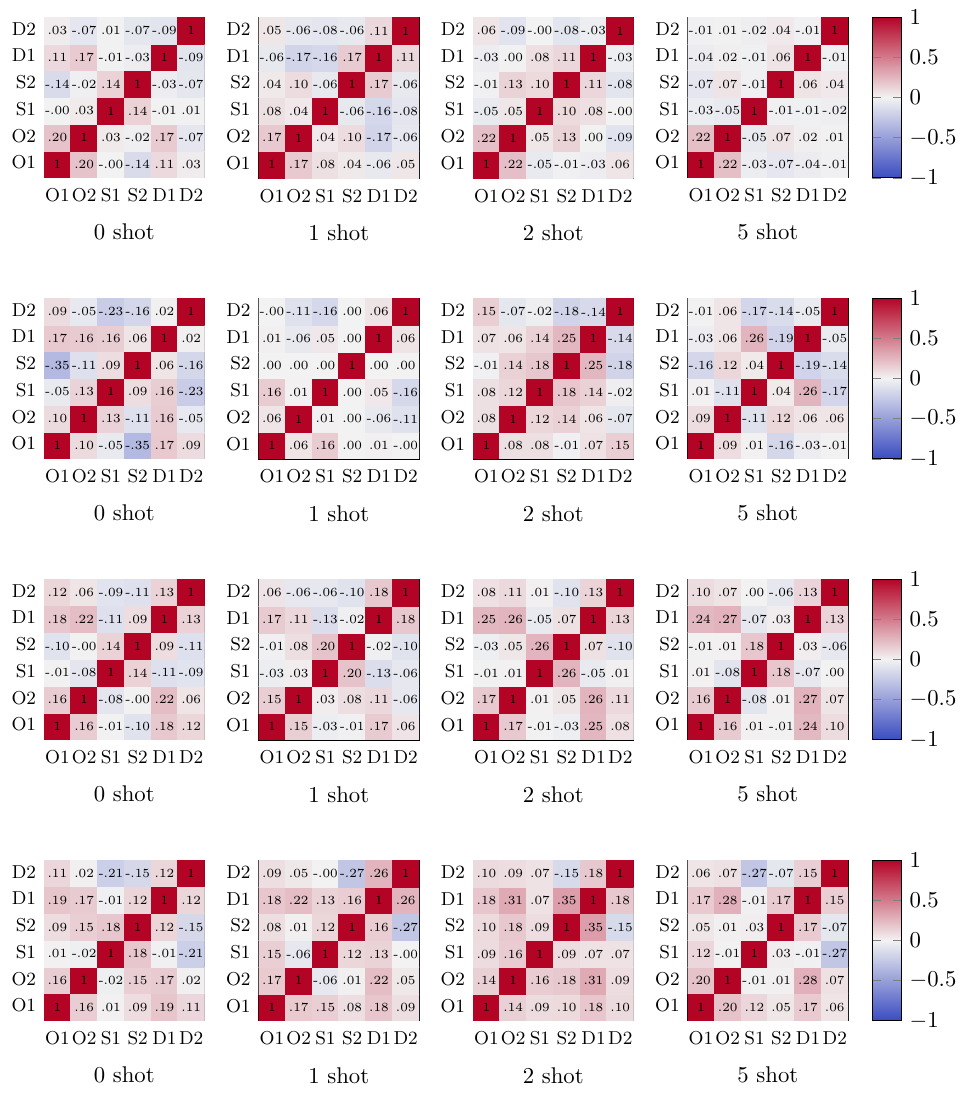}
  \caption{Correlation matrices across the six evaluation tasks—for Qwen (7B) and Qwen (72B). For each model, four panels correspond to prompt settings (0/1/2/5-shot). Rows and columns appear in the fixed order [O1, O2, S1, S2, D1, D2]; each cell reports the Pearson $\phi$ correlation between the binary correctness indicators of the two tasks, computed over the set of images common to both tasks (diagonal entries = 1). Colors encode correlation on a fixed range $[-1,1]$ (blue $\to$ negative, white $\approx 0$, red $\to$ positive), with the numeric value overlaid in each cell. } 
  \label{fig:app-corr-qwen-total}
\end{figure*}

Figure~\ref{fig:app-corr-qwen-total} reports Pearson correlations among all tasks across shot settings for Qwen 7B/72B under both all and high-risk splits.

\subsection{All-Scenario Evaluation}\label{app:eval-intern-excluded}
Tables~\ref{tab:app-eval-intern-excluded-1} and \ref{tab:app-eval-intern-excluded-2} list per-shot scores for all models.

\subsection{Similar-Scene Pair Robustness}\label{app:similar-scene}
We report single-image accuracy (Decision-1) and the joint accuracy that both images in a similar pair are answered correctly (Decision-1 (both)). Under an independence assumption, the expected baseline for the joint accuracy equals the squared single-image accuracy. Blue superscript stars indicate statistically significant drops of the joint accuracy below that squared baseline (one star for $|z|\in(1.65,1.96]$, two stars for $|z|>1.96$; two-sided normal approximation).

\subsection{Analyzer Robustness Across Models and Risk Strata}\label{app:analyzer-robustness}

To further assess analyzer stability, we evaluate 480 reasoning outputs using the same metrics as Table~\ref{tab:exp-match}.
This set extends the original 180 outputs with 300 newly annotated outputs, while keeping the reasoning format and analyzer prompt fixed.
The outputs are evenly split across Qwen2.5-VL-72B-Instruct, InternVL2.5-78B-MPO, and GPT-4.1, with 160 outputs per model and an 80/80 high-risk/non-high-risk split.
Table~\ref{tab:app-analyzer-robustness} shows similar results across model sources and risk strata, suggesting stable analyzer behavior under model and scenario-risk shifts.
Together with the representative examples in Appendix~\ref{app:exp-taxonomy}, this analysis supports using the analyzer for scalable auxiliary error-mode labeling.
\noindent
\begin{table}[t]
\centering
\small
\caption{Decision-1 accuracy on single images and on similar-scene pairs.}
\begin{tabular}{lccc}
\toprule
Model & Size & Decision-1 & Decision-1 (both) \\
\midrule
Gemma & 4B & 37.50 & \phantom{0}5.36\textcolor{blue}{$^{*\phantom{*}}$} \\
Gemma & 27B & 42.86 & 10.71 \\
Llama & 11B & 32.14 & \phantom{0}1.79\textcolor{blue}{$^{**}$} \\
Llama & 90B & 32.14 & 12.50 \\
Llava & 72B & \textbf{\underline{48.21}} & \textbf{\underline{21.43}} \\
Llava & 7B & 44.64 & \phantom{0}0.00\textcolor{blue}{$^{**}$} \\
Phi & 6B & 43.75 & \phantom{0}1.79\textcolor{blue}{$^{**}$} \\
Qwen & 7B & 39.29 & \phantom{0}5.36\textcolor{blue}{$^{*\phantom{*}}$} \\
Qwen & 72B & 44.64 & 10.71 \\
\bottomrule
\end{tabular}
\end{table}

\FloatBarrier
\clearpage
\begin{table*}[p]
\centering
\caption{Evaluation on All Scenarios}
\label{tab:app-eval-intern-excluded-1}
\setlength{\tabcolsep}{6pt}\renewcommand{\arraystretch}{1.05}\small
\begin{tabular}{lcccccccccc}
\toprule
Model & Size & Open & Shots & Object-1 & Object-2 & Scene-1 & Scene-2 & Decision-1 & Decision-2 & Avg Score \\
\midrule
Gemini-2.5-Pro & - & \xmark  & 0& 62.90 & \textbf{\underline{74.35}} & 76.73 & 25.61 & 49.47 & 55.11 & 57.36 \\
 &  &   & 1& 65.04 & 71.45 & 76.28 & 30.96 & 55.11 & \textbf{\underline{58.78}} & 59.60 \\
 &  &   & 2& 62.14 & 71.76 & 81.18 & 39.12 & 51.30 & 45.80 & 58.55 \\
 &  &   & 5& 63.05 & 72.98 & 80.29 & 38.68 & 51.45 & 53.44 & 59.98 \\
GPT-4.1 & - & \xmark  & 0& 67.48 & 73.74 & 92.54 & 46.44 & \textbf{\underline{56.95}} & 51.91 & \textbf{\underline{64.84}} \\
 &  &   & 1& 68.09 & 72.52 & 92.76 & 45.36 & 54.05 & 46.87 & 63.28 \\
 &  &   & 2& \textbf{\underline{69.16}} & 73.89 & \textbf{\underline{93.21}} & 45.36 & 54.81 & 46.87 & 63.88 \\
 &  &   & 5& 67.94 & 73.59 & 92.76 & 43.65 & 55.42 & 47.33 & 63.45 \\
\midrule
Gemma & 4B & \cmark  & 0& 49.86 & 50.07 & 27.34 & \phantom{0}0.89 & 45.78 & 30.66 & 34.10 \\
 &  &   & 1& 50.24 & 49.93 & 34.17 & 21.25 & 46.85 & 30.81 & 38.87 \\
 &  &   & 2& 50.99 & 52.26 & \phantom{0}7.28 & 17.14 & 45.64 & 37.68 & 35.17 \\
 &  &   & 5& 51.52 & 52.55 & 12.04 & 18.70 & 44.97 & 46.14 & 37.65 \\
Gemma & 27B & \cmark  & 0& 64.46 & 63.55 & 63.45 & \phantom{0}7.20 & 50.40 & 31.65 & 46.78 \\
 &  &   & 1& 64.15 & 63.10 & 64.56 & \phantom{0}8.17 & 50.55 & 30.74 & 46.88 \\
 &  &   & 2& 64.22 & 63.62 & 64.94 & 10.11 & 50.02 & 30.43 & 47.22 \\
 &  &   & 5& 64.67 & 62.95 & 64.71 & 12.73 & 50.10 & 30.13 & 47.55 \\
Llama & 11B & \cmark  & 0& 58.67 & 53.45 & 87.00 & 18.37 & 41.84 & 37.99 & 49.55 \\
 &  &   & 1& 45.86 & 46.65 & 22.29 & \phantom{0}0.07 & 34.12 & 39.18 & 31.36 \\
 &  &   & 2& 23.54 & 49.36 & 26.67 & \phantom{0}2.73 & 31.62 & 39.37 & 28.88 \\
 &  &   & 5& 40.59 & 42.78 & 20.95 & \phantom{0}2.92 & 26.94 & 37.40 & 28.60 \\
Llama & 90B & \cmark  & 0& 51.39 & 63.48 & 91.75 & 36.78 & 45.14 & 49.55 & 56.35 \\
 &  &   & 1& 58.62 & 61.50 & 66.64 & 21.57 & 41.16 & 45.64 & 49.19 \\
 &  &   & 2& 56.61 & 65.58 & 83.06 & 30.79 & 37.48 & 43.05 & 52.76 \\
 &  &   & 5& 59.62 & 64.83 & 83.13 & 35.42 & 40.57 & 47.28 & 55.14 \\
Llava & 7B & \cmark  & 0& 26.23 & 52.95 & 87.15 & 32.49 & 41.40 & 29.87 & 45.01 \\
 &  &   & 1& 26.59 & 47.59 & 72.51 & \phantom{0}7.31 & 34.34 & 42.17 & 38.42 \\
 &  &   & 2& 25.18 & 46.76 & 82.76 & 10.08 & 46.24 & 35.04 & 41.01 \\
 &  &   & 5& 25.69 & 50.19 & 85.88 & 16.19 & 49.90 & 29.20 & 42.84 \\
Llava & 72B & \cmark  & 0& 45.88 & 66.72 & 85.66 & 35.42 & 51.60 & 34.89 & 53.36 \\
 &  &   & 1& 55.79 & 64.82 & 82.62 & 22.71 & 44.38 & 41.97 & 52.05 \\
 &  &   & 2& 47.17 & 66.04 & 86.41 & 35.86 & 45.84 & 34.21 & 52.59 \\
 &  &   & 5& 55.53 & 66.04 & 84.70 & 34.22 & 52.23 & 34.28 & 54.50 \\
Phi & 6B & \cmark  & 0& 60.23 & 63.10 & 91.01 & 21.67 & 44.58 & 30.72 & 51.89 \\
 &  &   & 1& 62.71 & 53.80 & 82.09 & 13.35 & 39.37 & 39.07 & 48.40 \\
 &  &   & 2& 53.52 & 60.91 & 90.12 & 18.25 & 46.25 & 39.34 & 51.40 \\
 &  &   & 5& 62.02 & 59.70 & 85.51 & 24.15 & 34.34 & 46.09 & 51.97 \\
Qwen & 7B & \cmark  & 0& 65.58 & 59.34 & 65.16 & 33.60 & 40.87 & 35.48 & 50.00 \\
 &  &   & 1& 59.03 & 61.69 & 41.38 & 14.07 & 45.28 & 36.02 & 42.91 \\
 &  &   & 2& 60.97 & 62.20 & 53.34 & 22.66 & 42.22 & 35.56 & 46.16 \\
 &  &   & 5& 60.37 & 62.36 & 54.16 & 30.04 & 43.81 & 32.45 & 47.20 \\
Qwen & 72B & \cmark  & 0& 68.06 & 68.24 & 77.19 & 37.62 & 54.69 & 38.20 & 57.33 \\
 &  &   & 1& 68.53 & 68.76 & 77.05 & 45.07 & 54.17 & 42.31 & 59.31 \\
 &  &   & 2& 68.67 & 68.69 & 86.85 & 47.82 & 52.87 & 36.62 & 60.25 \\
 &  &   & 5& 68.67 & 68.38 & 87.59 & \textbf{\underline{48.98}} & 53.47 & 38.13 & 60.87 \\
\bottomrule
\end{tabular}
\end{table*}
\clearpage
\begin{table*}[p]
\centering
\caption{Evaluation on High-Risk Scenarios}
\label{tab:app-eval-intern-excluded-2}
\setlength{\tabcolsep}{6pt}\renewcommand{\arraystretch}{1.05}\small
\begin{tabular}{lcccccccccc}
\toprule
Model & Size & Open & Shots & Object-1 & Object-2 & Scene-1 & Scene-2 & Decision-1 & Decision-2 & Avg Score \\
\midrule
Gemini-2.5-Pro & - & \xmark  & 0& 77.40 & \textbf{\underline{70.62}} & 75.00 & 23.68 & 63.84 & \textbf{\underline{77.40}} & 64.66 \\
 &  &   & 1& 70.62 & 67.23 & 80.26 & 26.32 & 65.54 & \textbf{\underline{77.40}} & 64.56 \\
 &  &   & 2& 73.45 & 64.41 & 81.58 & 36.18 & 55.93 & 60.45 & 62.00 \\
 &  &   & 5& 71.75 & 65.54 & 84.21 & 30.48 & 61.58 & 64.97 & 63.09 \\
GPT-4.1 & - & \xmark  & 0& 80.23 & 68.93 & 92.76 & 48.90 & \textbf{\underline{71.75}} & 68.36 & \textbf{\underline{71.82}} \\
 &  &   & 1& 80.23 & 65.54 & 93.42 & 46.49 & 59.32 & 58.76 & 67.29 \\
 &  &   & 2& \textbf{\underline{83.05}} & 66.10 & 93.42 & 45.18 & 62.15 & 58.19 & 68.02 \\
 &  &   & 5& 81.92 & 66.10 & \textbf{\underline{94.74}} & 45.83 & 63.84 & 59.89 & 68.72 \\
\midrule
Gemma & 4B & \cmark  & 0& 67.62 & 37.05 & 24.14 & \phantom{0}0.00 & 55.88 & 36.34 & 36.84 \\
 &  &   & 1& 67.93 & 37.33 & 32.70 & 13.35 & 59.23 & 34.66 & 40.87 \\
 &  &   & 2& 67.90 & 41.52 & \phantom{0}7.21 & 11.68 & 59.57 & 50.28 & 39.69 \\
 &  &   & 5& 68.78 & 43.67 & 12.66 & 13.35 & 60.11 & 65.68 & 44.04 \\
Gemma & 27B & \cmark  & 0& 67.74 & 56.11 & 64.37 & \phantom{0}2.64 & 58.78 & 38.43 & 48.01 \\
 &  &   & 1& 67.99 & 56.67 & 59.71 & \phantom{0}8.52 & 58.19 & 38.23 & 48.22 \\
 &  &   & 2& 68.89 & 56.73 & 62.26 & 10.28 & 58.19 & 35.36 & 48.62 \\
 &  &   & 5& 68.61 & 56.73 & 60.15 & 12.91 & 57.26 & 35.67 & 48.56 \\
Llama & 11B & \cmark  & 0& 63.37 & 42.08 & 86.82 & 16.13 & 37.70 & 51.04 & 49.52 \\
 &  &   & 1& 52.50 & 40.49 & 23.09 & \phantom{0}0.00 & 65.34 & 54.30 & 39.29 \\
 &  &   & 2& 24.97 & 42.63 & 29.30 & \phantom{0}4.80 & 54.58 & 52.73 & 34.84 \\
 &  &   & 5& 44.00 & 36.49 & 28.47 & \phantom{0}5.33 & 51.15 & 54.98 & 36.74 \\
Llama & 90B & \cmark  & 0& 53.04 & 54.33 & 90.25 & 28.83 & 59.71 & 69.26 & 59.24 \\
 &  &   & 1& 61.43 & 52.14 & 71.24 & 21.36 & 66.58 & 63.82 & 56.10 \\
 &  &   & 2& 58.45 & 58.81 & 88.93 & 25.14 & 60.70 & 58.98 & 58.50 \\
 &  &   & 5& 61.29 & 55.88 & 88.58 & 29.31 & 65.73 & 63.77 & 60.76 \\
Llava & 7B & \cmark  & 0& \phantom{0}9.37 & 48.37 & 86.02 & 19.78 & 21.37 & 27.37 & 35.38 \\
 &  &   & 1& 10.64 & 36.32 & 83.13 & \phantom{0}2.34 & 65.34 & 61.82 & 43.26 \\
 &  &   & 2& \phantom{0}7.86 & 39.05 & 81.72 & \phantom{0}1.61 & 39.08 & 50.87 & 36.70 \\
 &  &   & 5& \phantom{0}7.91 & 43.78 & 85.99 & 11.01 & 50.25 & 27.99 & 37.82 \\
Llava & 72B & \cmark  & 0& 42.57 & 60.19 & 86.42 & 27.37 & 52.28 & 50.56 & 53.23 \\
 &  &   & 1& 51.75 & 57.26 & 78.19 & 17.99 & 62.89 & 62.16 & 55.04 \\
 &  &   & 2& 43.41 & 59.32 & 87.70 & 29.04 & 63.25 & 50.11 & 55.47 \\
 &  &   & 5& 51.72 & 58.39 & 86.86 & 27.45 & 60.47 & 49.27 & 55.69 \\
Phi & 6B & \cmark  & 0& 55.89 & 51.88 & 91.13 & 19.87 & 26.36 & 35.48 & 46.77 \\
 &  &   & 1& 61.63 & 42.46 & 86.11 & 12.10 & 64.19 & 53.07 & 53.26 \\
 &  &   & 2& 46.40 & 49.69 & 90.69 & 16.31 & 53.40 & 55.21 & 51.95 \\
 &  &   & 5& 55.18 & 47.72 & 85.11 & 23.16 & 64.78 & 65.14 & 56.85 \\
Qwen & 7B & \cmark  & 0& 59.57 & 49.97 & 68.55 & 40.27 & 33.42 & 44.63 & 49.40 \\
 &  &   & 1& 50.87 & 51.83 & 34.68 & 11.12 & 58.33 & 45.02 & 41.98 \\
 &  &   & 2& 54.28 & 51.40 & 48.69 & 15.74 & 44.40 & 43.39 & 42.98 \\
 &  &   & 5& 51.41 & 55.21 & 48.37 & 22.83 & 45.84 & 39.53 & 43.86 \\
Qwen & 72B & \cmark  & 0& 66.69 & 59.01 & 84.09 & 36.78 & 57.06 & 45.87 & 58.25 \\
 &  &   & 1& 67.80 & 58.19 & 76.16 & 45.91 & 60.19 & 51.74 & 60.00 \\
 &  &   & 2& 67.82 & 59.31 & 89.10 & \textbf{\underline{51.34}} & 53.07 & 40.66 & 60.22 \\
 &  &   & 5& 67.09 & 58.28 & 89.19 & 47.21 & 55.21 & 40.46 & 59.57 \\
\bottomrule
\end{tabular}
\end{table*}

\end{document}